\documentclass[sn-mathphys]{sn-jnl}

\usepackage{multirow}
\jyear{2021}%
\theoremstyle{thmstyleone}%
\theoremstyle{thmstyletwo}%

\theoremstyle{thmstylethree}%

\usepackage{fancyhdr}

\begin{document}

\title[Hossain et al.]{Biomembrane-based Memcapacitive Reservoir Computing System for Energy Efficient Temporal Data Processing}

\author[1]{\fnm{Md Razuan} \sur{Hossain}}\email{mhossai3@go.olemiss.edu}
\equalcont{These authors contributed equally to this work.}

\author[2]{\fnm{Ahmed S.} \sur{Mohamed}}\email{asm6015@psu.edu}
\equalcont{These authors contributed equally to this work.}

\author[2]{\fnm{Nicholas X.} \sur{Armendarez}}\email{nxa5323@psu.edu}

\author*[2]{\fnm{Joseph S.} \sur{Najem}}\email{jsn5211@psu.edu}

\author*[1]{\fnm{Md Sakib} \sur{Hasan}}\email{mhasan5@olemiss.edu}

\affil[1]{\orgdiv{Department of Electrical and Computer Engineering}, \orgname{University of Mississippi}, \orgaddress{\city{Oxford}, \state{ Mississippi}, \country{USA}}}

\affil[2]{\orgdiv{Department of Mechanical Engineering}, \orgname{The Pennsylvania State University}, \orgaddress{ \city{University Park}, \state{Pennsylvania}, \country{USA}}}

\abstract{Reservoir computing is a highly efficient machine learning framework for processing temporal data by extracting features from the input signal and mapping them into higher dimensional spaces. Physical reservoir layers have been realized using spintronic oscillators, atomic switch networks, silicon photonic modules, ferroelectric transistors, and volatile memristors. However, these devices are intrinsically energy-dissipative due to their resistive nature, which leads to increased power consumption. Therefore, capacitive memory devices can provide a more energy-efficient approach. Here, we leverage volatile biomembrane-based memcapacitors that closely mimic certain short-term synaptic plasticity functions as reservoirs to solve classification tasks and analyze time-series data in simulation and experimentally. Our system achieves a 99.6\% accuracy rate for spoken digit classification and a normalized mean square error of \unboldmath{$7.81\times10^{-4}$} in a second-order non-linear regression task. Furthermore, to showcase the device's real-time temporal data processing capability, we achieve a 100\% accuracy for a real-time epilepsy detection problem from an inputted electroencephalography (EEG) signal. Most importantly, we demonstrate that each memcapacitor consumes an average of 41.5 \unboldmath{$fJ$} of energy per spike, regardless of the selected input voltage pulse width, while maintaining an average power of 415 \unboldmath{$fW$} for a pulse width of 100 \unboldmath{$ms$}. These values are orders of magnitude lower than those achieved by state-of-the-art memristors used as reservoirs. Lastly, we believe the biocompatible, soft nature of our memcapacitor makes it highly suitable for computing and signal-processing applications in biological environments.}



\maketitle
\section{Introduction}\label{sec1}

\par Temporal data processing and time series prediction have recently gained increasing interest due to their ubiquitous utility in various fields such as speech recognition \cite{Graves2013SpeechNetworks, Sak2015FastRecognition}, language modeling, text generation \cite{Mikolov2010RecurrentModel, Wiseman2018LearningGeneration}, trend forecasting \cite{Salman2016WeatherTechniques}, traffic forecasting \cite{Li2017DiffusionForecasting}, and financial forecasting \cite{Selvin2017StockModel, Rout2017ForecastingApproach}. Temporal series prediction requires recurrent neural network (RNN) paradigms capable of history-dependent, multilayered input mapping to an output layer \cite{Alzubaidi2021ReviewDirections}. This means that the output is influenced by both the current and the prior inputs as well as the network states. Therefore, to achieve history-based computing, nodes in the hidden layers of conventional RNNs are recurrently or cyclically connected to themselves \cite{Alzubaidi2021ReviewDirections}. These cyclical connections increase the RNNs' computational cost and complexity of training as cyclic dependencies can exhibit bifurcations, and thereby non-convergence in training \cite{Doya1992BifurcationsNetworks, Lukosevicius2012ReservoirTrends}. To circumvent the training bottleneck, the concept of reservoir computing (RC) was independently introduced by Jaeger \textit{et al.} \cite{Jaeger2001TheNetworks} and Maass \textit{et al.} \cite{Maass2002Real-timePerturbations} in their RNN models, namely, the echo state network and the liquid state machines, respectively. 
\par
\begin{figure}
    \centering
    \includegraphics[width=4.5 in]{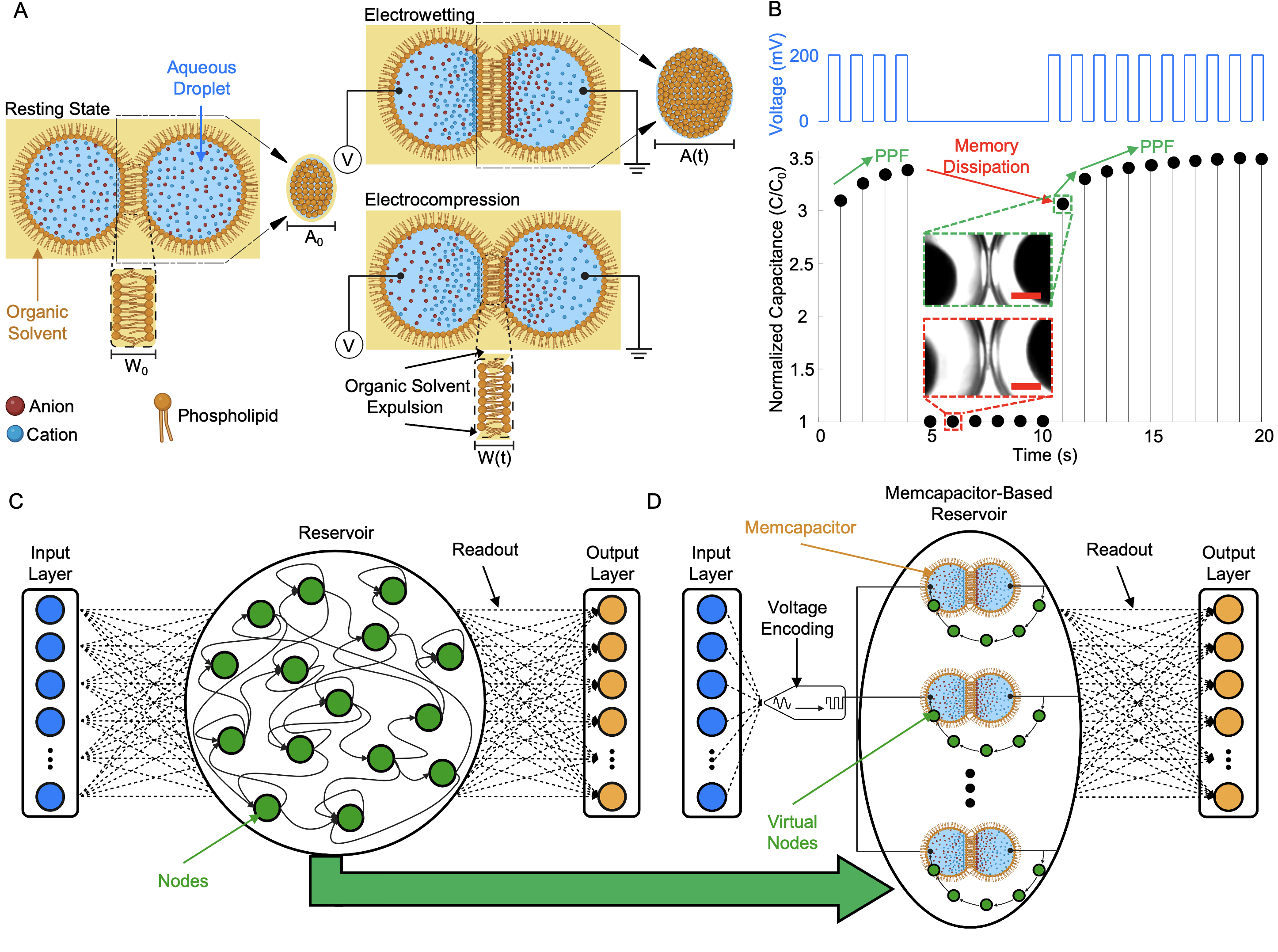}
    \caption{A memcapacitive reservoir computing scheme. \textbf{A)} A schematic describing a biomembrane-based memcapacitor in the resting state and in response to voltage. In response to voltage, this device exhibits hysteretic changes in membrane area (i.e., electrowetting) and thickness (i.e., electrocompression) which give rise to dynamical memcapacitance \cite{Najem2019DynamicalMembranes}. \textbf{B)} In response to a train of pulses, the memcapacitor outputs an increased relative capacitance from pulse to pulse (i.e., paired-pulse facilitation) due to its volatile memory. The inset enclosed in green-dashed displays an inverted microscope-obtained bottom-view of the memcapacitor exhibiting an almost two-fold increase in interfacial area compared to the interfacial area at rest (the inset enclosed in dashed red).The scale bars in red in both insets are equivalent to 200 $\mu m$ \textbf{C)} A representation of a conventional reservoir computing system entailing an input layer, the reservoir, a readout layer, and an output layer. \textbf{D)} An equivalent system can be built by replacing the reservoir layer with memcapacitors as reservoirs. The random spatial nodes in a conventional reservoir in panel \textbf{C} are replaced by serial temporal virtual nodes.}
    \label{RC}
\end{figure}

 RC is a brain-inspired emerging machine learning architecture \cite{Rabinovich2008Neuroscience.Processing}. As described in Figure \ref{RC}C, an RC incorporates an input layer that feeds the input signal to an RNN of fixed random weights called the reservoir \cite{Jaeger2001TheNetworks}. The reservoir, in most cases, nonlinearly maps the input signal to a high-dimensional state space that dynamically evolves with the time-varying input \cite{Jaeger2001TheNetworks}. The reservoir states, corresponding to all training inputs, then get projected to the output layer via the “memory-less” (i.e., time-independent) readout layer \cite{Jaeger2001TheNetworks}. Unlike conventional RNNs which necessitate weight training between every two subsequent layers \cite{Alzubaidi2021ReviewDirections}, the training of RC systems is constrained to the readout layer \cite{Jaeger2001TheNetworks}, which can be done, only once, using linear or logistic regression \cite{Verstraeten2007AnMethods, Lukosevicius2009ReservoirTraining}. Limiting the number of training layers to only one static readout layer offers a drastic reduction in the overall computational cost of the network when compared to conventional RNNs \cite{Jaeger2001TheNetworks}. This nonlinear mapping from low-dimensional to high-dimensional space increases input separability, thereby facilitating classification to different output classes \cite{Cover1965GeometricalRecognition}.
\par In practice, RC systems were initially implemented and studied in silico with marked temporal signal prediction accuracies \cite{Oztuik2007AnalysisNetworks, Schrauwen2007AnImplementations, Lukosevicius2012ANetworks}. However, soon after Appeltant \textit{et al.} \cite{Appeltant2011InformationSystem} theorized the equivalence of any dynamically-rich time-delay system to a dynamical reservoir, a spurt in more efficient hardware realizations of the reservoir occurred \cite{Tanaka2019RecentReview, Dunham2020PhysicalPerspective}. These hardware implementations included spintronic oscillators \cite{Torrejon2017NeuromorphicOscillators,Jiang2019PhysicalProcessing}, atomic switch networks \cite{Sillin2013AComputing, Lilak2021SpokenNetworks}, silicon photonic modules \cite{Vandoorne2014ExperimentalChip, VanDerSande2017AdvancesComputing}, ferroelectric transistors \cite{Toprasertpong2022ReservoirTransistor, Duong2023DynamicProcessing}, and most notably memristors \cite{Du2017ReservoirProcessing, Moon2019TemporalSystem, Midya2019ReservoirMemristors, Zhu2020MemristorAnalysis, Hossain2021ReservoirMemristor, Zhong2021DynamicProcessing, Cao2022EmergingComputing}. Memristors, short for “memory resistors”, are two-terminal, state-dependent resistive elements that co-locate volatile memory  and, in many cases, complex nonlinear dynamics \cite{Chua1971MemristorTheElement}, hence the prevalence of memristor-based RC systems in the literature. Memristor-based RC systems achieved remarkable performance in various applications including, yet not limited to, hand-written \cite{Du2017ReservoirProcessing, Midya2019ReservoirMemristors} and spoken digit recognition \cite{Moon2019TemporalSystem, Zhong2021DynamicProcessing}, chaotic time series prediction \cite{Moon2019TemporalSystem, Zhong2021DynamicProcessing, Hossain2022MemristorPrediction}, and real-time neural firing pattern classification \cite{Zhu2020MemristorAnalysis}.
\par In theory, any dynamical system with sufficient short-term memory and nonlinearity can act as a reservoir \cite{Appeltant2011InformationSystem}. Therefore, reservoirs can be generalized to encompass a whole class of nonlinear electronics called memelements. The term “memelements” was coined by Chua \cite{Chua1980DeviceElements} to refer to any passive circuit element with a response dependent on input history, such as memristors, memcapacitor, meminductors, as well as other higher order memelements \cite{Abdelouahab2014Memfractance:Memory, Biolek2016EveryTheorem}. Memcapacitors, in particular, have garnered attention in computing applications \cite{Pershin2014MemcapacitiveNetworks, Demasius2021Energy-efficientComputing} that required very low power consumption due to their energy-storing nature as opposed to the energy-dissipative nature of memristors, in addition to the pulse-width-independent energy per spike consumption. Although various research groups have proposed theoretical models for memcapacitive devices \cite{DiVentra2009CircuitMeminductors, Martinez-Rincon2010Solid-stateCapacitance, Pershin2014MemcapacitiveNetworks, Mohamed2015ModelingJunctions, Khan2016MonolayerDevice}, only a few physical implementations have been demonstrated \cite{You2016AnFunctions, Wang2018CapacitiveNeuro-transistors, Zheng2019ArtificialCharacteristics, Kwon2020CapacitiveApplications}. In terms of in silico neuromorphic computing applications, memcapacitor models \cite{Najem2019DynamicalMembranes, Mohamed2015ModelingJunctions, Biolek2013ReliableMeminductors} have been employed in the simulation of memcapacitive networks, where the memcapacitors serve only as variable weights between the network's nodes \cite{DatTran2017MemcapacitiveComputing, DatTran2020DeepNetwork, Tran2023Multi-taskingNetworks, Tran2019HierarchicalArchitecture}. Such architectures utilize nodal activation functions for the nonlinear transformation, as the memcapacitor itself does not perform the nonlinear mapping. To the best of our knowledge, there have been no reports of memcapacitor-based RC system architectures where the memcapacitor nonlinearly maps an input to a higher dimensional feature space with fading memory, thereby fully replacing the network-based reservoir layer. For such a memcapacitor-based RC system, it is imperative that the memcapacitor exhibits both nonlinearity with respect to input excitation and volatility (i.e., the fading memory property) \cite{Maass2002Real-timePerturbations, Du2017ReservoirProcessing}. The utility of physical memcapacitors in neuromorphic computing applications has been limited to artificial neural networks due to the constraint of nonvolatile memcapacitors \cite{Demasius2021Energy-efficientComputing}. Additionally, to the best of our knowledge, there have been no reports on the physical implementation of a memcapacitor-based reservoir computing system \cite{Wang2018CapacitiveNeuro-transistors, Zheng2019ArtificialCharacteristics}.

\par Here, we experimentally realize a memcapacitor-based RC system that leverages two-terminal, volatile, scalable, memcapacitors, first introduced by Najem \textit{et al.} \cite{Najem2019DynamicalMembranes}, as reservoirs that fully replace conventional network-based RC architectures due to their intrinsic nonlinearity and state volatility. The memcapacitor consists of a synthetic lipid bilayer formed between two lipid-encased aqueous droplets submerged in an oil phase \cite{Najem2019DynamicalMembranes}. At the interface of both droplets, an elliptical, planar lipid bilayer ($\sim 100$ $\mu m $ in radius) spontaneously forms with a highly insulating ($>100$ $M\Omega \cdot cm^2$) core consisting of a mixture of hydrophobic lipid tails and residual entrapped oil (Figure \ref{RC}A). Upon transmembrane voltage application, the ionically charged lipid bilayer manifests geometrical changes due to electrowetting (EW) and electrocompression (EC) (Figure \ref{RC}A), leading to an increase in bilayer area and a decrease in the hydrophobic, respectively (see Supplementary Section 1 and Figure S1-S4 for more details). The bilayer exhibits dynamical voltage-controlled capacitance with paired-pulse-facilitation (PPF) (Figure \ref{RC}B) via geometric reconfigurability of its interfacial area and hydrophobic thickness \cite{Najem2019DynamicalMembranes}, enabling the high-dimensional temporal transformation with minimal power and energy consumption. First, we demonstrate the device's computational quality by conducting spoken-digits recognition as well as predicting a second-order dynamical time series and compare the achieved accuracies one-to-one with another memristor-based RC report \cite{Du2017ReservoirProcessing, Moon2019TemporalSystem}. Then, taking advantage of the device's biological synapse-like time-scale ($\sim$10\textsuperscript{2} $ms$ \cite{Zucker2003Short-TermPlasticity}),  we detect epilepsy from an electroencephalography (EEG) signal to demonstrate the device's real-time temporal processing. Furthermore, for completion, we solve an IRIS dataset classification problem to confirm that the device's short-term dynamics can solve a static classification problem (Supplementary Information Section 5). Finally, we present the device’s power and energy per spike consumption and compare them with the consumption of other state-of-the-art memristors which were deployed as reservoirs \cite{Du2017ReservoirProcessing, Najem2018MemristiveMimics, Midya2019ReservoirMemristors, Zhong2021DynamicProcessing, Zhu2020MemristorAnalysis, Maraj2023SensoryPerformance}.

\section{Results}\label{sec2}


\subsection{Memcapacitor-based Reservoir Computing Architecture}\label{sec2_2}

\par The structure of Reservoir Computing (RC) is based on a three-layered architecture comprising an input layer, a reservoir, and an output layer \cite{Lukosevicius2009ReservoirTraining} as shown in Figure \ref{RC}C. The input layer is responsible for encoding and distributing the input data to the reservoir, a large, fixed, and sparsely connected network of non-linear dynamic nodes \cite{Jaeger2001TheNetworks}. These nodes, often referred to as neurons or perceptrons, are interconnected through random and fixed weights, enabling the reservoir to serve as a rich temporal memory \cite{Jaeger2001TheNetworks, Maass2002Real-timePerturbations}. The output layer consolidates the high-dimensional reservoir states into a meaningful output, with only the connections between the reservoir and output layer being adjusted during training \cite{Schrauwen2007AnImplementations, Gallicchio2017DeepAnalysis}. This unique structure allows RC to efficiently handle many temporal tasks, outperforming traditional RNNs in terms of computational complexity and training time and solidifying its position as a powerful and versatile approach to processing time series data \cite{Schrauwen2007AnImplementations, Butcher2013ReservoirAnalysis}. 
\par Taking advantage of the memcapacitor’s second-order nonlinear dynamics in which electrowetting and electrocompression operate at two distinct time scales \cite{Najem2019DynamicalMembranes}, short-term memory (Figure \ref{RC}B and Supplementary Figures S2-S4), and innate stochasticity (Supplementary Figure S5), we replaced the conventional RNN-based RC with a memcapacitor-based RC, as shown in Figures \ref{RC}C and \ref{RC}D. Unlike memristor-based reservoirs, where the resistance or conductance of the memristor is commonly representative of the reservoir state \cite{Du2017ReservoirProcessing, Yang2013MemristiveComputing, Moon2019TemporalSystem, Zhong2021DynamicProcessing}, in memcapacitor-based reservoirs, the reservoir state is reflected by capacitance, or, as more often used in this work, capacitance normalized to its resting capacitance state ($C/C_0$). It is worth noting that there are nontrivial dissimilitudes between a conventional RC and a memcapacitor-based RC in terms of their computing architecture. A conventional RC typically incorporates hundreds, often thousands, of interconnected cyclic nodes \cite{Tanaka2019RecentReview}. Contrarily, a memcapacitor-based reservoir exploits the memcapacitor’s inherent time-delay aspect (i.e., memory) to account for the nodes' recursion \cite{Appeltant2011InformationSystem}. Furthermore, the nonlinearity of a single memcapacitor can compensate for the nonlinearity offered by a number of interconnected nodes, depending on the device’s degree of nonlinearity and quality of fading memory (i.e., volatility). Accordingly, as hypothesized by Appeltant \textit{et al.} \cite{Appeltant2011InformationSystem}, a single dynamical system with sufficient nonlinearity and time-delay, such as our memcapacitor, can act as a full reservoir, which promotes practical implementations of physical RCs. It should be noted that it has been mathematically proven that a reservoir state cannot concurrently render high degrees of nonlinearity and memory capacity because nonlinearity degrades memory \cite{Dambre2012InformationSystems, Inubushi2017ReservoirTrade-off}. 

\par For a memcapacitor-based RC with a few memcapacitors connected in parallel, as depicted in Figure \ref{RC}D, an effective mapping of the input signal to a high-dimensional state space can be accomplished by means of virtual nodes \cite{Appeltant2011InformationSystem}. When first introduced \cite{Appeltant2011InformationSystem}, virtual nodes were adapted by sampling many points along the reservoir’s time-delayed response to every input time step. In other words, every input value sent to the reservoir is sampled-and-held via a voltage encoder (Figure \ref{second_order}A), yielding time-varying reservoir states that span the holding time. Each sampled point along the time-delayed response sequence can be deemed an independent node and state-space dimension, hence the name "virtual node" (Figure \ref{RC}D) \cite{Appeltant2011InformationSystem}. However, for high-frequency sampling, consecutive nodes are close enough to be considered linearly dependent, which is redundant for high-dimensional mapping; thus, it is more functional to choose an adequately-spaced sample from the obtained array to designate as the reservoir’s virtual nodes. Interestingly, for input sequences with negligible variations, as seen in cochleogram channels fully populated with 0-bits (channels 3-20 in Figure \ref{audio}), to avoid redundantly large feature spaces, it is more advisable to assign a virtual node once every few inputs rather than more than once per input. To be comprehensive, in practical implementations, one can choose to elect as many or as few virtual nodes as demanded by the task of interest from the full array of measured reservoir states. Furthermore, for some applications, applying a form of post-processing on elected virtual nodes before transmitting them to the readout layer can also improve the overall system's performance. 
\par In this study, we discuss the employment of the memcapacitor-based reservoir for solving three distinct problems, where the input encoding and the feature space definition for each problem were executed differently. We start with a benchmark spoken-digit classification problem, where the input signal is binary (either a \lq{0}\rq{} or a \lq{1}\rq{}) and temporally history-dependent (Figure \ref{audio}). For this problem, a virtual node was elected for every five equally spaced inputs. Then, we discuss how we used the memcapacitor-based reservoir for predicting a second-order regression problem, where the input signal is random, continuous (non-binary), and history-dependent. Unlike the spoken-digit problem, the input signal was encoded in 10 different timescales to effectively increase the dimensionality of the reservoir. We then present an epilepsy detection problem using an EEG input signal as a real-time temporal signal processing problem. Leveraging the device's biological fading memory (100 $ms$) \cite{Zucker2003Short-TermPlasticity}, we incorporate a feature modification post-process that integrates 60 short-timescale (5.8 $ms$) features into one long-timescale (348 $ms$) cumulative feature (i.e., virtual node) to be passed to the readout layer. As a supplementary problem (Supplementary Section 5), we solve a static classification problem, namely the IRIS dataset problem, where the input is history-independent to demonstrate the device's exemplary performance across tasks with dissimilar inputs.



\subsection{Spoken Digit Classification}\label{sec2_3}

\par
\begin{figure}
    \centering
    \includegraphics[width=4.5 in]{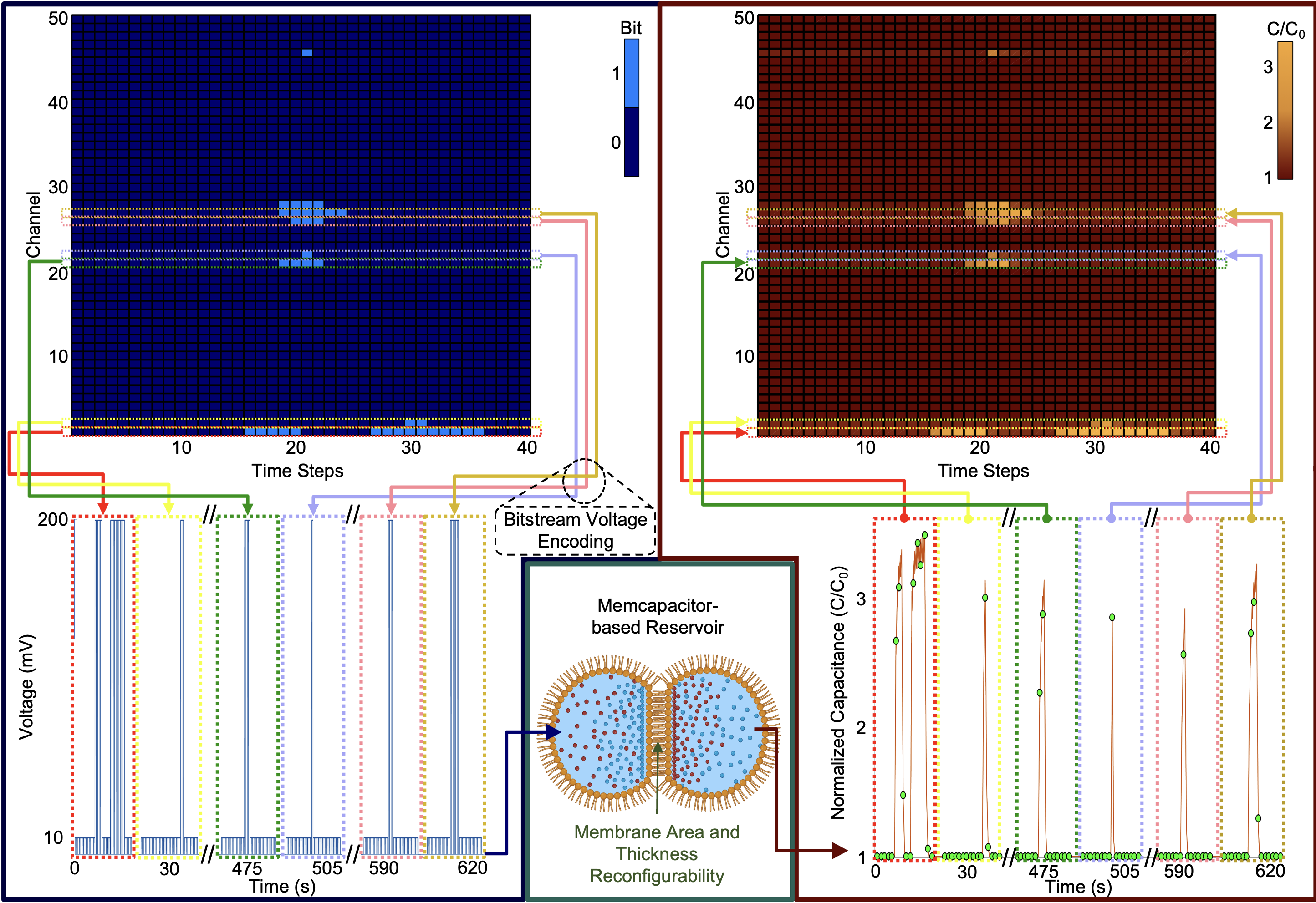}
    \caption{The encoding process flow of the Spoken Digit classification problem. The binary 2-D cochleogram (top-left corner) represents neural spike trains in different human cochlear channels. Each input was converted into voltage pulses (bottom-left corner), where the 10-$mV$ and 200-$mV$ pulses correspond to a \lq{0}\rq{} (resting neuron) bit and a \lq{1}\rq{} (firing neuron) bit, respectively. The input voltage train was fed to the memcapacitor-based reservoir, where the normalized capacitance response is recorded (bottom-right corner). The dynamic normalized capacitance was then mapped to a 2-D matrix (top-right corner) and, for every channel or row, one virtual node is selected for every 5 timesteps as depicted by the green circles on the capacitance plot (bottom-right corner).}
    \label{audio}
\end{figure}

 Here, we demonstrate the performance of our memcapacitor-based RC system for a benchmark speech recognition task. The dataset we used, NIST TI46, comprises binary cochleograms of isolated spoken digit waveforms, 0-9 in spoken English (Supplementary Figure S6). The dataset was provided to us as a courtesy by the authors of Moon \textit{et al.} \cite{Moon2019TemporalSystem}. It consists of 500 binary 2D cochleograms, with 450 used for training and the remaining 50 for testing. Each cochleogram, an example of spoken digit \lq{6}\rq{}  is shown in Figure \ref{audio}, digitally represents the response of 50 human cochlear channels to sound waves captured over 40 timesteps. For each channel in the vertical axis, a 40-bit binary sequence renders a neuron firing event along 40 timesteps, where the light blue area denotes a firing event or 1-bit while the dark blue area signifies no firing or 0-bit for the corresponding time step. Subsequently, 1-bit and 0-bit time steps were encoded respectively as 200-$mV$ and 10-$mV$ square pulses (500-$ms$ pulse width), as shown in the voltage vs. time plot at the bottom-left corner of Figure \ref{audio}. 

Our analysis of the 500 datasets revealed 25000 channels in total, with 795 being distinct. To streamline the experimental setup, we routed the 795 unique bitstreams (Supplementary Figure S7) that comprise all 25000 channels to a single memcapacitor. The corresponding capacitance responses were then estimated from the recorded current and normalized to the resting state at every time step using the method described in the Methods section. The upper right inset in Figure \ref{audio} depicts a 2-D heat-map of the normalized capacitance response (Supplementary Figure S8) resulting from inputting the example spoken digit \lq{6}\rq{} cochleogram in the upper left inset of Figure \ref{audio}. As seen in Figure \ref{audio}, the shown cochleogram indicated a predominance of 0 bits, suggesting that the reservoir state was mostly inactive, which was also true for all 500 cochleograms comprising the training and testing dataset. Accordingly, selecting a virtual node after each time step led to an unreasonably large feature space (31800 dimensions for 795 channels  and 40 timesteps), as explained in Section \ref{sec2_2}. Such a large feature space could result in states' linear dependence, and therefore information redundancy. To overcome this problem, we divided the input sequence into $n$ equal intervals (every 5 time steps in length, where capacitance data were recorded at the end of each interval, yielding a total of 8 virtual nodes per channel. These elected virtual nodes are represented by the green circles in the bottom-right inset of Figure \ref{audio}. The reservoir is comprised of a total of 50$n$ virtual nodes, given that there are 50 frequency channels, each containing $n$ virtual nodes. These modes are used in a one-vs-all readout layer (50$n$ × 10) to classify among ten digits. The fitcecoc function in MATLAB was utilized with a logistic regression learner for training the readout network.  It is important to note that while only one memcapacitor is required to process all channels, we could have used multiple memcapacitors in parallel to speed up the encoding process with the same accuracy and energy consumption.



\par
\begin{figure}
    \centering
    \includegraphics[width=4.5 in]{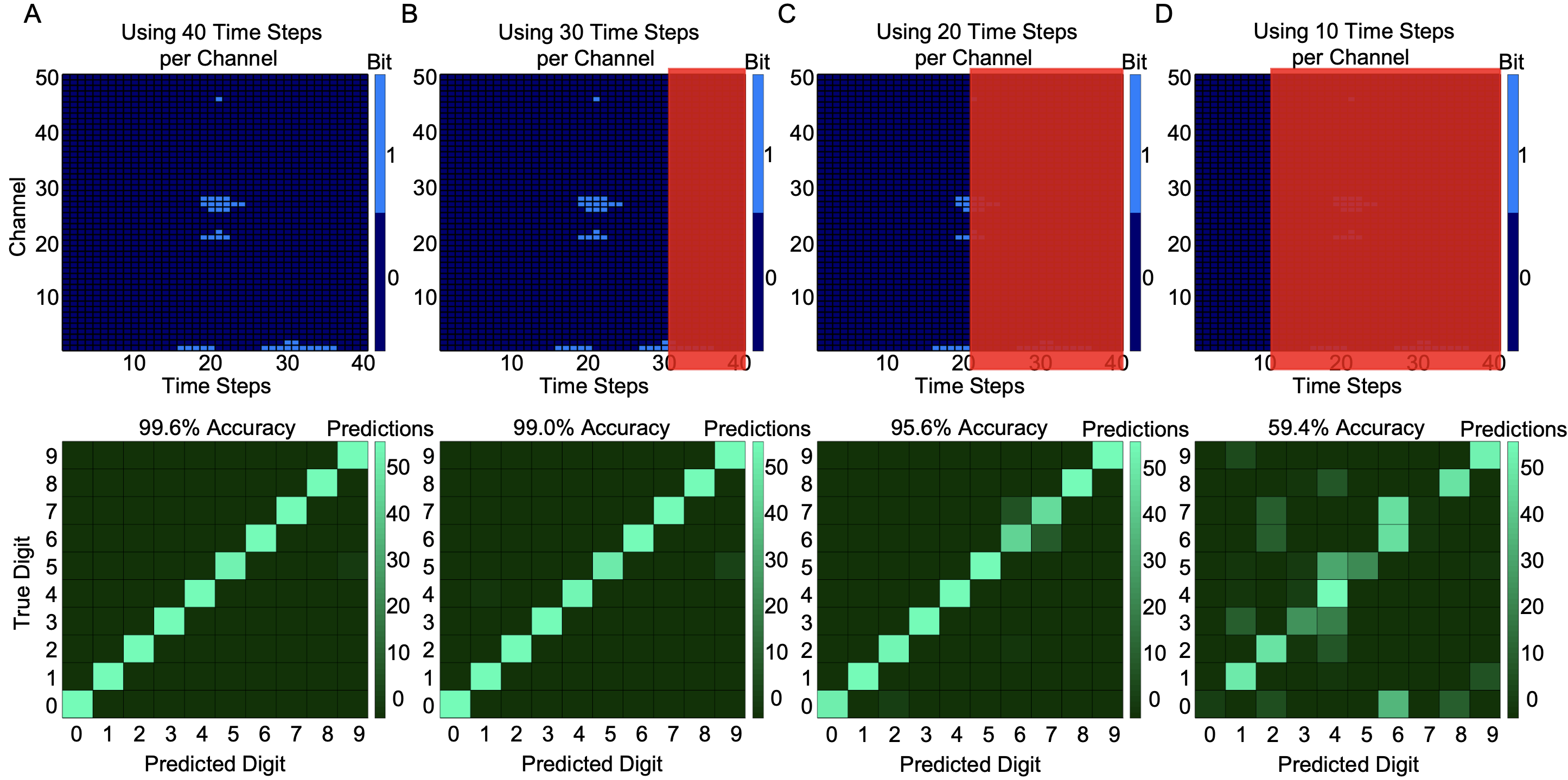}
    \caption{Prediction results for the Spoken Digit classification problem using a memcapacitor-based RC system. \textbf{A)} Confusion Matrix for spoken Digits using all timesteps in every channel (100\% of the utterance). The overall testing dataset accuracy is 99.6\%. \textbf{B)} Confusion Matrix for spoken Digits using 30 out of 40 timesteps per channel (75\% of the utterance). The overall testing dataset accuracy is 99.0\%. \textbf{C)} Confusion Matrix for spoken Digits using all time steps 20 out of 40 timesteps per channel (50\% of the utterance). The overall testing dataset accuracy is 95.6\%. \textbf{D)} Confusion Matrix for spoken Digits using only 10 out of 40 timesteps per channel (25\% of the utterance) The overall testing dataset accuracy is 59.4\%.}
    \label{MNISTAudioResults}
\end{figure}

 In summary, as an average of 10 runs, our memcapacitive RC system attained a $99.6\%$ success rate in experiments and $100\%$ in simulations for all inputs. Further, we have also evaluated the performance of our system at $75\%$, $50\%$, and $25\%$ of the input completion, meaning that our RC system predicts the spoken digit before the utterance is fully delivered. The results are summarized in confusion matrices plotted in Figure \ref{MNISTAudioResults}. In addition, Table \ref{pblm1} presents a comparison of recent work \cite{Du2017ReservoirProcessing} for experimental accuracy and the reservoir energy (Supplementary Figure S9).

\begin{table}[]
 \caption{Recognition rate and energy comparison of the Spoken Digit classification in an experimental framework. (Note: the comparison of energy has solely been conducted on the reservoir of the test dataset.)}
 \centering
\begin{tabular}{|c|c|c|c|c|}
\hline
\textbf{\begin{tabular}[c]{@{}c@{}}Time\\ Steps\end{tabular}} &
  \textbf{\begin{tabular}[c]{@{}c@{}}This \\ Work\end{tabular}} &
  \textbf{\begin{tabular}[c]{@{}c@{}}This \\ Work\end{tabular}} &
  \textbf{\begin{tabular}[c]{@{}c@{}}Moon \\ \textit{et al.} \cite{Moon2019TemporalSystem}\end{tabular}} &
  \textbf{\begin{tabular}[c]{@{}c@{}}Moon \\ \textit{et al.} \cite{Moon2019TemporalSystem}\end{tabular}} \\ \hline
   & \textbf{Accuracy (\%)} & \textbf{Energy (J)} & \textbf{Accuracy (\%)} & \textbf{Energy (J)} \\ \hline
10 & 59.4                     & 4.22$\times10^{-9}$             & 57.8                   &  2.68$\times10^{-6}$             \\ \hline
25 & 97.0                     & 1.05$\times10^{-7}$             & 98.2                   & 1.41$\times10^{-5}$             \\ \hline
40 & 99.6                     & 1.19$\times10^{-7}$             & 99.2                   & 2.55$\times10^{-5}$             \\ \hline
\end{tabular}
 \label{pblm1}
\end{table}

\subsection{Solving a second-order nonlinear dynamic task}\label{sec2_4}


Nonlinear dynamical systems are mathematical models that describe the complex behavior of natural and engineered systems that are characterized by time-dependent interactions among their constituent elements \cite{Aguirre2009ModelingReview}. The study of these nonlinear systems has proven invaluable in elucidating a broad spectrum of phenomena in various disciplines, such as fluid dynamics \cite{Alomari2013ApproximateProblem}, population biology \cite{Varga2008ApplicationsBiology}, climatic systems \cite{Palmer1999APrediction} etc. The governing equations for second-order nonlinear dynamical systems encompass time derivatives of the second order. Some prominent examples of second-order nonlinear dynamics within these domains include electrical system converters \cite{Li2013NonlinearSystem} as well as  damping properties in mechanical systems \cite{Ruderman2021OptimalSystems}, among others \cite{XuAControl, Vo2021AApplications, Mei2015DistributedSystems}. In this section, we use a memcapacitive reservoir to predict a second-order dynamic nonlinear transfer function \cite{Du2017ReservoirProcessing}. The transfer function is described as follows:

\begin{equation}
y(t)=0.4y(t-1)+0.4y(t-1)y(t-2)+0.6u^3(t)+0.1
\label{eq:SE}  
\end{equation}
\newline
\noindent The output signal, $y(t)$, depends on the present input, $u(t)$, as well as the previous two inputs, $y(t-1)$ and $y(t-2)$ (i.e., a time lag of two-time steps) as shown in Eq. (\ref{eq:SE}). In this study, we trained the memcapacitor-based RC system to map a random input onto a higher-dimensional space, thereby enabling the generation of an accurate second-order dynamic nonlinear transfer function output from the input after training without prior knowledge of the underlying mathematical relationship between input and output.

\begin{figure}
    \centering
    \includegraphics[width=4.5 in]{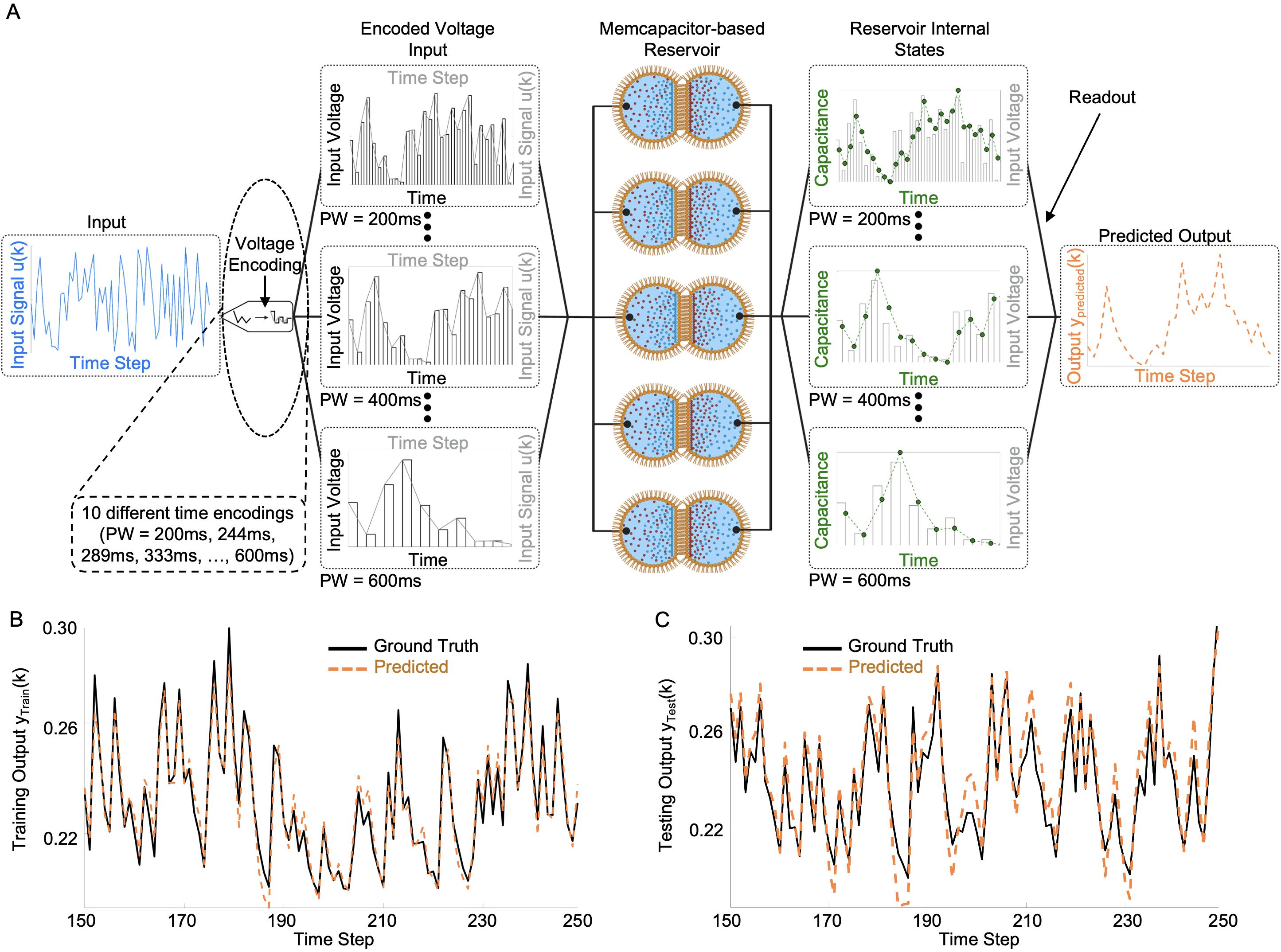}
    \caption{Solving a second-order nonlinear dynamic task using a memcapacitor-based RC system. \textbf{A)} The process flow of solving second-order dynamic tasks. The random input signal gets encoded (i.e., sampled and held) with 10 different pulse widths (PW) ranging from 200 $ms$ to 600 $ms$. The encoded voltage was then fed to five memcapacitors and the corresponding capacitance is measured at the end of every voltage pulse for all 10 different encodings. We assigned a virtual node to every capacitance measured, resulting in 50 virtual nodes per time step (5 memcapacitors by 10 time encodings). Subsequently, a 50-by-300 state matrix is passed to the readout layer for output prediction using linear regression. \textbf{B)} Experimentally obtained output prediction compared to the ground truth output using the training dataset. The estimated NMSE is $5.75\times10^{-4}$. \textbf{C)} Experimentally obtained output prediction compared to the ground truth output using the testing dataset. The estimated NMSE is $7.81\times10^{-4}$.}
    \label{second_order}
\end{figure}

Figure \ref{second_order}A presents the schematic of the RC system developed for solving a second-order nonlinear dynamic task. We initially channel the input signal into the voltage encoder, which converts the input into voltage pulses. The voltage pulses are then directed to the memcapacitive reservoir, generating different reservoir states. These states are subsequently employed by the readout function to derive the anticipated output.


We chose a random input signal sequence within the range of 0 to 0.5 and transformed it into a voltage amplitude between 50 $mV$ and 200 $mV$. We employ a random sequence of 300 time frames as inputs for training the memcapacitor-based RC system. The reservoir is composed of 50 memcapacitive virtual nodes within the employment of 5 physical memcapacitive devices. For each memcapacitor, the input voltages are provided via pulse streams of $50\%$ duty cycle with 10 different time frame widths (equally spaced between 200 $ms$ and 600 $ms$) applied to each of the 5 memcapacitors in the reservoir throughout the experiment. It is worth noting that there exists a slight variation across the five memcapacitors due to inherent slight differences in bilayer area and hydrophobic thickness (denoted $A_0$ and $W_0$, respectively, in Figure \ref{RC}A). We found that including 5 devices with slightly varying properties enhances reservoir performance, as inherent device-to-device variations contribute to more separable reservoir outputs. Comparable performance improvement was observed for inputs with 10 distinct time frames or pulse widths (PW). We found that having 50 reservoir states (comprising 5 devices, each receiving 10 input encodings) was optimal since further increasing the number of reservoir states increased the computational overhead without significantly improving the prediction accuracy. In this instance, the readout layer is a 50 × 1 feedforward layer and serves to convert the reservoir state into a single output. A simple linear regression model with gradient descent is utilized to train the weights in the readout layer. The process flow of solving the second-order dynamic task is demonstrated in Figure \ref{second_order}A.

Panels B and C in Figure \ref{second_order} present the graphical representation of the original and predicted signals for 100 time frames of training and testing data in the experiment, respectively. Notably, the readout function is not re-trained during the testing phase. The normalized mean square error (NMSE) values (refer to Supplementary Information Section 3.1) attained for training and testing data amount to $5.28\times10^{-4}$ and $6.32\times10^{-4}$ in simulation (Supplementary Figure S10), and $5.75\times10^{-4}$ and $7.81\times10^{-4}$ in experiment, respectively. Table \ref{pblm2} summarizes a comparison with recent work \cite{Du2017ReservoirProcessing}. Furthermore, it is important to note that solving this problem using a reservoir based on a conventional linear network yielded a larger NMSE than that of the memcapacitor-based reservoir (Supplementary Figure S11 and Supplementary Information Section 3.2). This highlights the indispensability of the intrinsic nonlinear dynamics of the memcapacitor device for higher dimensional transformation.





 



\begin{table}[]
\caption{Comparison table of solving second-order nonlinear dynamic system. (Note: Here, reservoir energy has been compared only for the test dataset.)}
\centering
\begin{tabular}{|c|c|c|c|c|c|}
\hline
\textbf{Work} &
  \textbf{\begin{tabular}[c]{@{}c@{}}Train\\ (NMSE)\end{tabular}} &
  \textbf{\begin{tabular}[c]{@{}c@{}}Test\\ (NMSE)\end{tabular}} &
  \textbf{\begin{tabular}[c]{@{}c@{}}Reservoir\\ States\end{tabular}} &
  \textbf{\begin{tabular}[c]{@{}c@{}}Reservoir\\ Energy ($J$)\end{tabular}} 
   \\ \hline
This Work &
  $5.75\times10^{-4}$ &
  $7.81\times10^{-4}$ &
  $50$ &
  $2.72\times10^{-8}$ \\ \hline
Du \textit{et al.} \cite{Du2017ReservoirProcessing}&
  $3.61\times10^{-3}$ &
  $3.13\times10^{-3}$ &
  $90$ &
  $3.34\times10^{-4}$\\ \hline
\end{tabular}
 \label{pblm2}
\end{table}


\newpage
\subsection{Real-time Epilepsy Detection from EEG Signal}\label{sec2_5}


Electroencephalogram (EEG) signals, which are time series data, serve as a valuable tool for examining abnormal brain activity occurring during seizure episodes, as these signals are captured from diverse areas within the brain \cite{Tzallas2009EpilepticAnalysis}. Representing temporal patterns of brain activation, EEG signals contain essential information about brain functionality \cite{Basar2012APathology}. Thus, by extracting pertinent features, researchers can gain a more profound understanding of the brain's underlying activities \cite{Reynolds2018ATasks}. The primary objective of this study is to demonstrate the effectiveness of our memcapacitor-based RC system capable to solve classification problems in real-time.


In this study, we have used a dataset from the University of Bonn, Germany \cite{Andrzejak2001IndicationsState}. This dataset is composed of five distinct classes of EEG signals, specifically F, N, S, Z, and O. Each class contains 100 EEG signals with a duration of 23.6 seconds, sampled at a frequency of 173.67 $Hz$ (5.8 $ms$) and exhibiting voltage amplitudes below and above of $\pm$1000 $\mu V$. Class S EEG signals are captured during seizure activity, while class Z EEG signals are obtained from healthy individuals \cite{Nishad2020ClassificationTransform}. The primary focus of this research is the real-time classification of class Z (healthy) and S (epileptic) EEG signals.

\begin{figure}
    \centering
    \includegraphics[width=4.5 in]{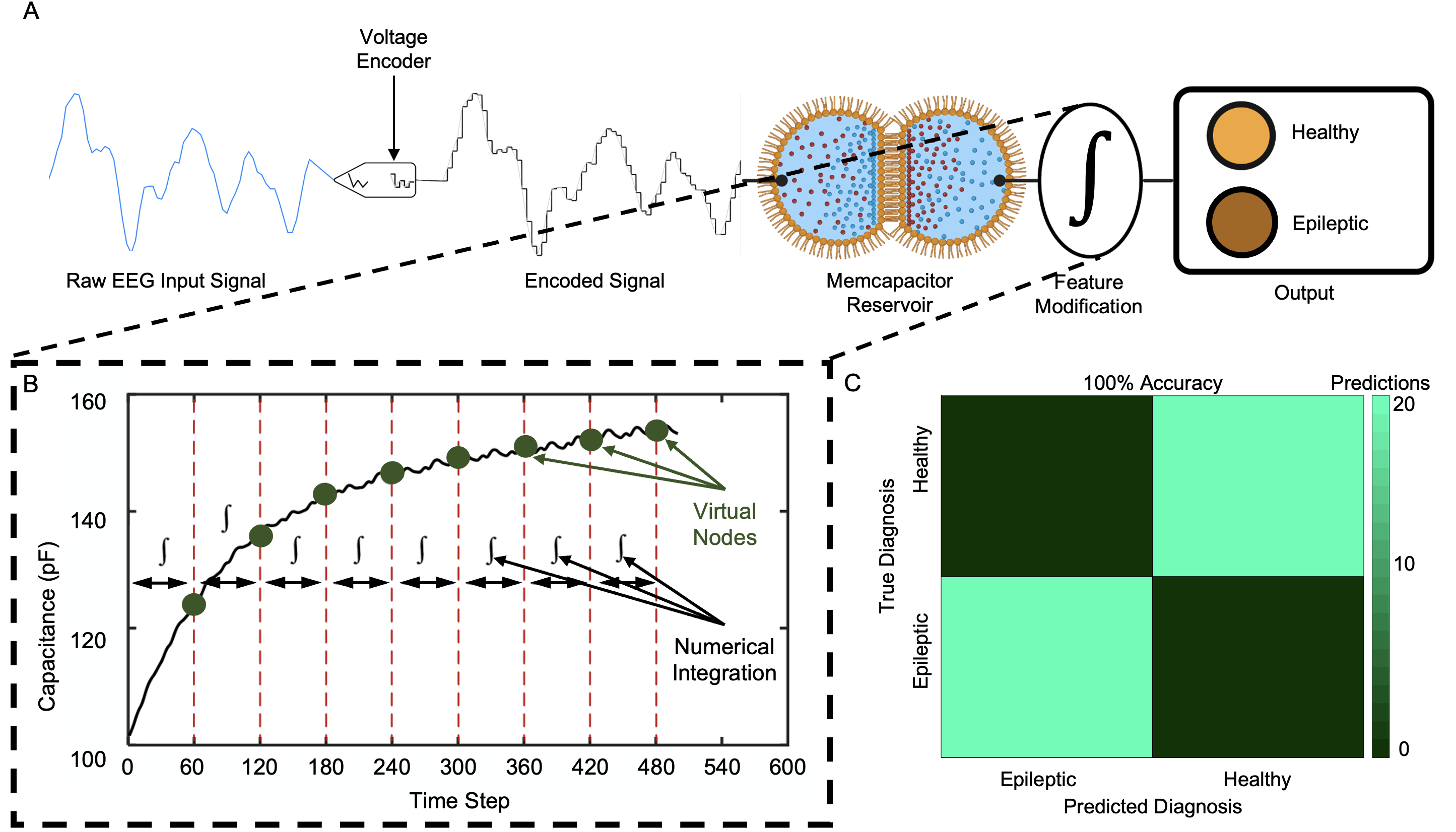}
    \caption{ Epilepsy detection task from an input EEG signal using a memcapacitor-based RC system. \textbf{A)} The process flow of EEG Dataset classification using a memcapacitor-based RC system. Note that the additional feature modification step precedes the readout. \textbf{B)} The feature modification post-mapping step numerically integrates the memcapacitor states every 60-time steps and combines these 60 features into one combined feature. \textbf{C)} A confusion matrix summarizing the prediction results for the EEG signal classification using the memcapacitor-based RC system. The results convey a 100\% testing accuracy.}
    \label{EEG}
\end{figure}

Figure \ref{EEG} presents the process flow for EEG dataset classification. In order to assess the classification performance of our system using minimal information from raw data, we took the absolute value of the signal and clipped it to the highest value of $300$  $\mu V$. The signal is then converted into a voltage pulse train ranging between 100 $mV$ and 200 $mV$, with a pulse width of approximately 6 ms. This pulse train is then introduced to a memcapacitive reservoir, which causes alterations in the reservoir's dynamics. To optimize feature size, a feature modification layer is added after the reservoir, where integration occurs.

For instance, each EEG data point consists of a voltage signal extending over 4097-time steps. We divided the entire capacitance value into 68 virtual nodes. As illustrated in Figure \ref{EEG}, we integrated the capacitance between virtual nodes rather than capturing it at the virtual node itself, recording the value at the $(60n)^{th}$ step (with $n = 1, 2, 3, ..., 68$). In this manner, the capacitance value for 60-time steps generates a single feature, leading to a total of 68 features for each data point, thereby enabling real-time classification. While the virtual node technique reduces feature sizes, it retains less past information. Conversely, the virtual node method with the integration technique preserves past information from previous steps, ultimately enhancing performance. In the output layer, logistic regression is performed. By employing the memcapacitor-based RC system, we achieved 100\% accuracy for both training and testing EEG data.

\subsection{Energy consumption of Memcapacitor-based RC}\label{sec2_6}

\begin{figure}
    \centering
    \includegraphics[width=3.5 in]{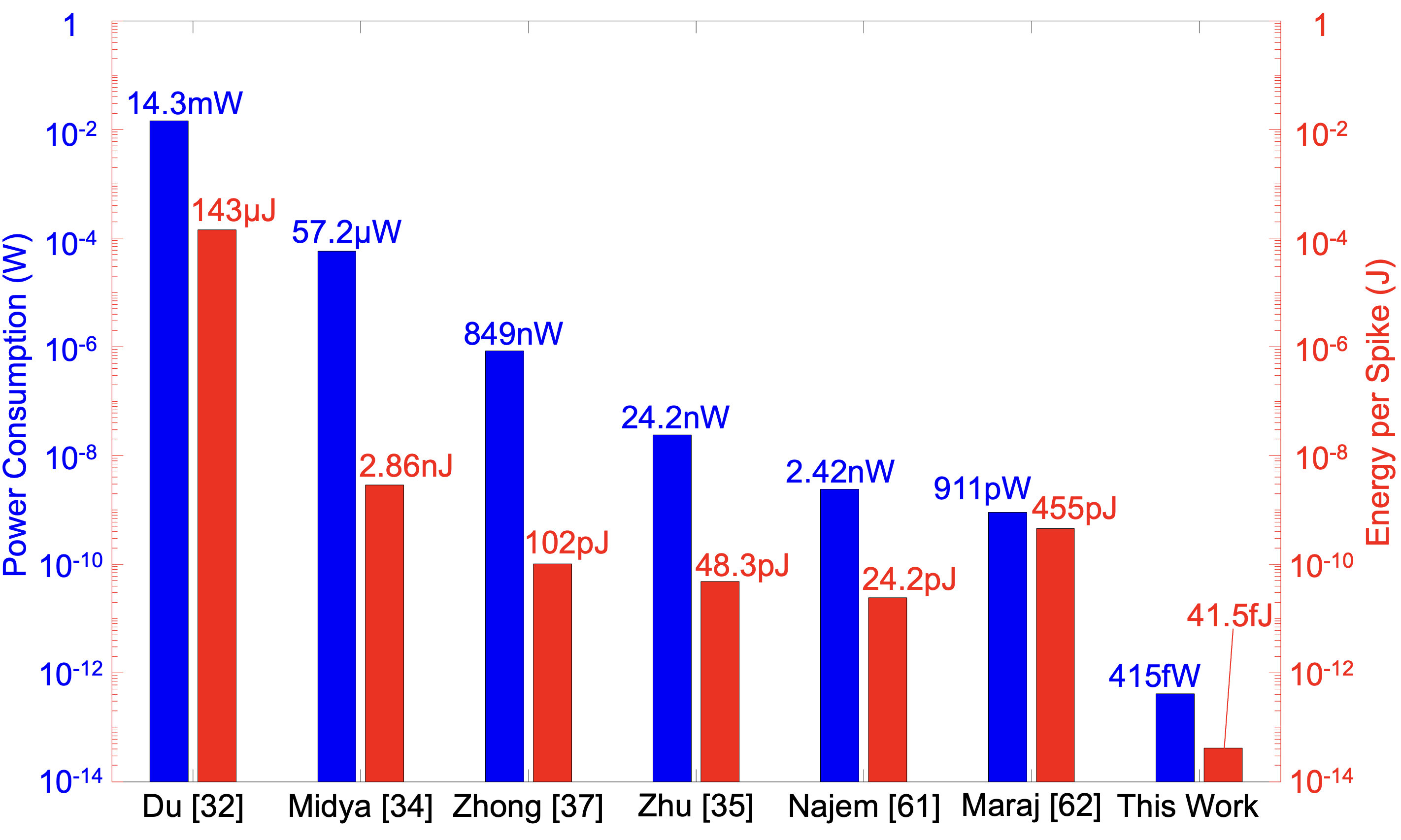}
    \caption{A comparison of power and energy per spike consumption is plotted on a log-scale, showcasing the significant difference in power and energy between the memcapacitor used in this study and state-of-the-art memristors employed as reservoirs. The memcapacitor exhibits orders of magnitude lower power and energy per spike consumption compared to the memristor \cite{Midya2019ReservoirMemristors, Du2017ReservoirProcessing, Najem2018MemristiveMimics, Zhu2020MemristorAnalysis, Zhong2021DynamicProcessing,Maraj2023SensoryPerformance}.}
    \label{energy}
\end{figure}

In order to evaluate the energy efficiency of our memcapacitor, we computed the mean power as well as energy consumed per spike via simulation and compared it to six state-of-the-art memristors that were previously employed for reservoir computing. (Du \textit{et al.} \cite{Du2017ReservoirProcessing}, Midya \textit{et al.} \cite{Midya2019ReservoirMemristors}, Zhu \textit{et al.} \cite{Zhu2020MemristorAnalysis}, Zhong \textit{et al.} \cite{Zhong2021DynamicProcessing}, Najem \textit{et al.} \cite{Najem2018MemristiveMimics}, Maraj \textit{et al.} \cite{Maraj2023SensoryPerformance}). We transmitted a uniform distribution random signal for each device model ranging from 0 to 1, spanning 1000 data points with appropriate voltage encoding parameters. That is, for every memristor device model, the random input signal was encoded into a voltage pulse train with a pulse width and voltage range that match the values reported in their corresponding articles. Then, the resulting unique 1000-pulse trains were fed to their corresponding devices, and the current responses were simulated via the reported device models. The point-by-point products of the simulated currents and input voltages were then averaged to yield power consumption, numerically integrated, and then averaged per pulse width basis to yield the mean energy consumption per spike. In contrast, the energy per spike calculation of the memcapacitor was done using the product of simulated capacitance and the square of the voltage difference for increasing voltage amplitudes only. Meaning that we only consider the energy involved in charging the memcapacitor and disregard the discharging energy. As a result, the memcapacitor's energy per spike and mean power are almost five and three orders of magnitude, respectively, lower than that of the most energy-efficient memristor \cite{Zhu2020MemristorAnalysis} included in this study. In addition, the energy per spike of the memcapacitor is independent of the chosen input voltage pulse width, in contrast to memristors and, more generically, energy-dissipative devices in which energy consumption scales with the voltage pulse width. This is due to the fundamental dependence of capacitive current on the dynamic voltage transience as opposed to the dependence of ohmic current on the static voltage magnitude and on-time. As a result, memcapacitors only consume dynamic energy during pulse transition and do not consume any static energy during the pulse duration making the energy consumption completely independent of the pulse width. In the context of real-time memristor-based reservoir computing, the task of interest dictates the input voltage pulse width, and therefore the energy required to solve the problem. For instance, to classify distinct neural activity patterns in real-time, Zhu \textit{et al.} \cite{Zhu2020MemristorAnalysis} had to select a pulse width of 2 $ms$ for their input voltage to mimic the spike time of a biological action potential even though their memristor is able to operate at shorter time scales (500 $\mu s$). This problem-related imposition of pulse width resulted in an increase in their energy consumption per spike from 50 $fJ$ to almost 10 $pJ$. Unlike memristor-based reservoirs, the energy consumption of real-time memcapacitor-based reservoir computing is solely dependent on the device's capacitance and encoded input voltage magnitude, which are both dictated by the device's properties. For our memcapacitor, the energy consumption is a function of capacitance, which is proportional to the bilayer area and, consequently, the droplets' sizes (Figure \ref{RC}A and Supplementary Figure S1). For this analysis, we chose 50 $nL$-sized droplets, which ultimately yield a mean capacitance of $\sim$20 $pF$ and a 41.5 $fJ$ of energy per spike and a mean power consumption of 415 $fW$ for a 100-$ms$ pulse width. A log-scale bar graph is indicated in Figure \ref{energy} as a visual summary of the results. 

\section{Discussion}\label{sec3}

In this study, we present a physical memcapacitive RC system that takes advantage of the inherent short-term memory capabilities of biomembrane-based memcapacitive devices to fully replace network-based RC systems. Our approach utilizes the concept of virtual nodes for prediction tasks in the context of spoken digit classification problems. Notably, our reservoir system demonstrates the ability to classify data even when faced with incomplete inputs. When tackling regression such as second-order nonlinear dynamic problems, our system can accurately predict the actual output, even in the absence of knowledge about the transfer function. To emphasize the importance of the fading memory property of the device for this specific issue, we compared our memcapacitive RC system with a conventional linear network. We observed a significant difference in prediction NMSE (see Supplementary Information Section 3.2), indicating the superiority of our approach. Furthermore, the versatility of our system is demonstrated through its ability to process not only temporal data but also static, real-time data classification. In this regard, we have successfully addressed the epilepsy detection problem. Additionally, we solved Iris dataset classification problem (Supplementary Information Section 5 and Figure S12) in simulation to demonstrate the device's capability in solving problems with a static input. Finally, we demonstrated, in simulation, that the energy per spike consumption of the memcapacitor is not only orders of magnitude less than state-of-the-art memristors but also is independent of the pulse width of the input voltage, which is of particular interest in real-time physical reservoir computing.

The memcapacitor-based RC system presented in this work is expected to pave the way for ongoing advancements targeting the optimization of network performance across a diverse array of applications, such as action recognition, prediction, and classification. This method holds particular promise for applications that prioritize network size and energy efficiency over rapid processing speeds. In our current implementation, the reservoir is constructed in hardware, while input and output layers reside within a simulation platform. As part of our future research endeavors, we aim to build the entire system in hardware, thereby addressing real-time application challenges more effectively. Additionally, we plan to incorporate algorithmic advancements, such as improving the input encoding and new memcapacitive reservoir architecture, into our work. These improvements in simulation and experimental contexts will establish the groundwork for the application of memcapacitive RC systems in machine learning tasks tailored to neuromorphic computing applications.

We note that while solid-state memristors have achieved notable integration into VLSI circuits and are pervasive in industrial and commercial applications, in many different classes of applications, it is more appropriate to implement computation directly with devices that compute with biological mechanisms. We underscore that the goal of such biomolecular-device-based computing is not to compete with electronic circuits in terms of computational speed or size. The main advantage of biomolecular-device-based computing systems is their environment of application along with energy efficiency.

A primary motivation behind our work is directly linked to the objective of physical reservoir computing, where the main aim is to harness the intrinsic computational capabilities of specific materials while possessing physical properties desirable for applications that are not directly connected to their computational purpose \cite{Tanaka2019RecentReview}. Previous research indicates the potential of this approach, particularly in edge computing \cite{Tanaka2019RecentReview, Zhu2020MemristorAnalysis} for biomedical and brain-computing interfaces, where the inherent attributes of CMOS-based devices may be less advantageous in the biological milieu. However, iontronic, biocompatible tools, such as our memcapacitor, are more likely to be suited. Though memcapacitors may not match the scalability of traditional CMOS-based systems, strides have been \cite{Schimel2021Pressure-drivenNetworks, Nguyen2016MicrofluidicBilayers, Nguyen2016HydrodynamicArrays, Nguyen2017AMembranes, Villar2013AMaterial, Alcinesio2020ControlledTissues, Alcinesio2022FunctionalNetworks, Graham2017High-ResolutionPrinting, Sarles2010PhysicalNetworks, Challita2018EncapsulatingOrganogel} and are still being made to advance their scalability. Such neuromorphic devices and architectures exhibit a profound potential to offer rich dynamics and computing paradigms that reduce the need for extensive nanoscale devices, as evident from our and past studies \cite{Zhong2021DynamicProcessing, Du2017ReservoirProcessing, Zhu2020MemristorAnalysis, Maraj2023SensoryPerformance}.

Despite scalability limitations compared to solid-state devices, by exploring and harnessing the unique properties of volatile memcapacitive devices, we seek to revolutionize the field of reservoir computing and contribute to the development of more efficient and versatile neuromorphic systems. As the demand for intelligent and energy-efficient computing solutions grows, the findings of this study hold significant potential for transforming a wide range of industries and applications. Ultimately, the memcapacitive RC system presented here is a crucial step forward in the pursuit of next-generation, energy-efficient neuromorphic computing systems.

\section{Methods}\label{sec4}
\subsection{Lipid Solutions Preparation and Membrane assembly}\label{method_1}
 An aqueous stock solution containing 500-$mM$ potassium chloride (KCl, Sigma), 10-$mM$ 3-morpholino propane-1-sulfonic acid (MOPS, Sigma) with a measured pH of 5.8, and 2-$mg/mL$ 1,2-Diphytanoyl-sn-glycero-3-phosphocholine (DPhPC, Avanti) liposomes in deionized water (18.2 $M\Omega$$\cdot$cm) were prepared and stored. To prepare this aqueous stock solution, 160 $\mu L$ of 25-$mg/mL$ DPhPC lipids dissolved in chloroform solution are acquired and evaporated under clean dry air, leaving 4 $mg$ of residual lipid cake at the bottom of a 4-$mL$ vial. The vial is then left under vacuum for a minimum of two hours and then hydrated with 2$mL$ of  500-$mM$ KCl, 10-$mM$ MOPS buffer solution, resulting in an aqueous solution containing 2-$mg/mL$ multilamellar DPhPC liposomes. To convert the multilamellar liposomes to unilamellar liposomes, the lipid solution is first subjected to 6 freeze/thaw cycles, and then extruded by forcing it, in 11 immediately successive passes, through a 100-$nm$-pore polycarbonate membranes (Whatman) using an Avanti Mini Extruder. Finally, the extruded solution is sonicated for 5 minutes and vortex mixed for 60 seconds \cite{Najem2019AssemblyMembranes, Taylor2015Heating-enabledExtract, Taylor2015DirectBilayer}. This solution can be stored for weeks at 4$^oC$ or directly used for experimentation. A micropipette was used to pipette two 200-$nL$ droplets from the prepared lipid stock solution onto two 125 $\mu m$-diameter, ball-end silver/silver chloride (Ag/AgCl 99.99\%, Goodfellow) wires submerged in a decane ($\geq$99\%, Sigma-Aldrich) oil-filled, transparent acrylic reservoir. Prior to droplet deposition, the wires were coated with 1\% agarose gel to avoid droplets detaching from the wires due to the decreasing surface tension associated with the lipid monolayer formation. 

Initially, the droplets were suspended on the wires free of contact with each other, the acrylic substrate, and the oil/air interface for 5-7 minutes, allowing for a packed lipid monolayer to form at the droplets’ water/oil interface. The monolayer formation was monitored visually via a 4x objective lens on an Olympus IX73 inverted microscope. Once the monolayers were formed, which was detected by observing the droplets leave the lens’ focal plane as they vertically sagged from the wires, the droplets were brought in contact with each other by moving the wires using 3-axis micro-manipulators to spontaneously form a bilayer at the contact interface. 

\subsection{Electrical Measurements Setup}\label{method_2} Prior to implementing any problem-specific transmembrane voltage signal, we ensured a successful interfacial bilayer formation by supplying a triangular 10-$Hz$, 10-$mV$ voltage signal to the electrodes via a Tektronix AFG31022 function generator. As a result of the membrane’s high insulation ($>$100 $M\Omega$$\cdot$cm) and capacitive interfacial area, a small 10-$Hz$ square ($\sim 20$ $pA$) current response is expected as an output from a non-leaking bilayer. To obtain dynamic and steady-state changes in capacitance as a function of voltage (Supplementary Figure S2, S3, and S5), a voltage waveform consisting of a 50-$mHz$, 150-$mV$ amplitude sinusoidal waveform superimposed on a 20-$Hz$, 10-$mV$ triangular waveform was supplied to the membrane. The slow-frequency component of the waveform drove the geometric reconfiguration of the membrane (i.e., electrowetting and electrocompression) while the fast-frequency component was used to obtain the capacitance magnitude at every semi-period. For the capacitance step response in Supplementary Figure S4, the low-frequency component was a 10-second, 150-$mV$ square wave. A custom MATLAB script (available upon request) was used to compute the capacitance at every semi-period by fitting the analytical solution of a parallel RC circuit current response to the measured current. Simultaneously, the bilayer area changes (Supplementary Figure S1) were monitored at 30 fps using a camera attached to the inverted microscope. The corresponding videos, obtained via Olympus CellSens software, were then post-processed using a custom MATLAB script (available upon request) to extract the bilayer’s interfacial minor axis radius, which was used to compute the interfacial area changes (Supplementary Figure S2) and subsequently, the bilayer hydrophobic thickness changes (Supplementary Figure S2).
A custom MATLAB script (available upon request) was used to control an NI 9264 voltage output module to send an arbitrary transmembrane voltage signal. The capacitance magnitude at every pulse was computed from the obtained current responses using a custom MATLAB script (available upon request). 
 \par All the current measurements were recorded and digitized at $50,000$ samples/second (to avoid capacitive-spike aliasing) using a patch-clamp amplifier Axopatch 200B and Digidata 1440A data acquisition system (Molecular Devices), respectively. All current recordings are conducted on an active vibration isolation table and under appropriate shielding, using a lab-made Faraday cage, to reduce the noise to less than 2 $pA$.
\par All model simulations and energy consumption calculations were implemented using a custom MATLAB script (available upon request).


\end{document}


\title[]{Supplementary Information}


\author[1]{\fnm{Md Razuan} \sur{Hossain}}\email{mhossai3@go.olemiss.edu}
\equalcont{These authors contributed equally to this work.}

\author[2]{\fnm{Ahmed S.} \sur{Mohamed}}\email{asm6015@psu.edu}
\equalcont{These authors contributed equally to this work.}

\author[2]{\fnm{Nicholas X.} \sur{Armendarez}}\email{nxa5323@psu.edu}

\author*[1]{\fnm{Joseph S.} \sur{Najem}}\email{jsn5211@psu.edu}

\author*[2]{\fnm{Md Sakib} \sur{Hasan}}\email{mhasan5@olemiss.edu}

\affil[1]{\orgdiv{Department of Electrical and Computer Engineering}, \orgname{University of Mississippi}, \orgaddress{, \city{Oxford}, \state{ Mississippi}, \country{USA}}}

\affil[2]{\orgdiv{Department of Mechanical Engineering}, \orgname{The Pennsylvania State University}, \orgaddress{ \city{State College}, \state{Pennsylvania}, \country{USA}}}



\maketitle
\newpage
\section{Short-term Memory and Nonlinearity in Memcapacitors}\label{sec1}

\par Memory capacitors, or memcapacitors in short, are two-terminal nonlinear energy storage elements that exhibit memory properties, whereby the magnitude of capacitance nonlinearly depends on one or more internal states and can be regulated based on present and past external stimulation. Like memristors, memcapacitors can be categorized as either nonvolatile if the memcapacitors’ states are maintained or volatile if their states are unmaintained upon removal of an electrical stimulus \cite{Yang2013MemristiveComputing} Herein, replicating a device developed by Najem \textit{et al.} \cite{Najem2019DynamicalMembranes} formerly, we constructed a lipid bilayer-based parallel-plate memcapacitor $(\sim 0.1-1)$ $\mu F \cdot cm^{-2}$ \cite{Sarles2010RegulatedSubstrates, Najem2018MemristiveMimics} by interfacing two lipid monolayer-encased aqueous droplets ($\sim 200$ $nL$ each) immersed in an oil phase. At the interface of the two droplets, an elliptical, planar lipid bilayer ($\sim 100$ $\mu m $ in radius) spontaneously forms with a highly insulative $(>100$ $M\Omega \cdot cm^2)$ core that is comprised of a mixture of hydrophobic lipid tails and residual entrapped oil (main text Figure 1A). Upon transmembrane voltage application, the ionically charged lipid bilayer manifests geometrical changes due to electrowetting (EW) and electrocompression (EC) (main text Figure 1A), i.e., bilayer area increase (Figure S1-S5) and hydrophobic thickness decrease, respectively. EW is mainly caused by the bilayer tension reduction \cite{Requena1975TheFilms, Taylor2015DirectBilayer} due to charge-induced electrostatic forces; meanwhile, the electrostatic force-driven entrapped oil expulsion is the main drive for EC \cite{Evans1975MechanicsMembranes, Najem2018MemristiveMimics}. Gradients with respect to time in the minor axis radius, $R(t)$, of the bilayer area $A(t)$ and thickness, $W(t)$, are modeled using the coupled state equations \cite{Najem2019DynamicalMembranes} below: 

\begin{equation}
\frac{dR(t)}{dt} =  \frac{1}{\zeta_{ew}}((\frac{a\epsilon\epsilon_{0}}{2W(t)}) v(t)^2- k_{ew}(R(t)-R_0))
\end{equation}	

\begin{equation}
\frac{dW(t)}{dt} =  \frac{1}{\zeta_{ec}}((\frac{-a\epsilon\epsilon_{0}\pi R(t)^2}{2W(t)^2}) v(t)^2+ k_{ec}(W_0-W(t)))
\end{equation}

where $a$ is the eccentricity of an ellipse, $\epsilon$ is the equivalent dielectric constant for the hydrophobic tails and residual oil mixture, $\epsilon_{0}$ is the permittivity of free space, $R_0$ ($m$) is the zero-volt, interfacial area minor axis radius, $W_0$ ($m$) is the zero-voltage hydrophobic thickness, $\zeta_{ew}$  and $k_{ew}$ are the EW effective damping ($Nsm^{-2}$) and stiffness ($Nm^{-2}$) coefficients in the tangential direction, respectively, and $\zeta_{ec}$ and $k_{ec}$ are the EC effective damping ($Nsm^{-1}$) and stiffness ($Nm^{-1}$) coefficients in the normal direction, respectively. Similar to a standard parallel-plate capacitor, the dynamic capacitance, $C(R(t), W(t))$, can be expressed as: 
				
\begin{equation}
C(R(t),W(t))= \frac{\varepsilon \varepsilon_0 A(t)}{W(t)} =  \frac{\varepsilon \varepsilon_0 (a \pi R(t)^2)}{W(t)}
\end{equation}

\par The aforementioned voltage-induced area and thickness changes (Supplementary Figure S2 and S3) correspond to an analog, nonlinear 2-3 times increase in capacitance (Main text Figure 1B and Supplementary Figure S3). In addition to the nonlinear dependence on absolute voltage, the corresponding capacitance variations exhibit short-term plasticity, particularly  paired-pulse facilitation (PPF) \cite{Lopez2001AFacilitation} (Main text Figure 1B and Supplementary Figures S2-S4). In Figure 1B in the main text, we show the memcapacitor’s monotonic increase in normalized capacitance to an initial train of four $200$ $mV$ pulses, insinuating short-term plasticity, where the memcapacitance was computed at the end of each pulse. Following the first four pulses, the device was left unstimulated for 6.5 seconds, which restored the initial capacitance state, conveying memory loss in the device. In fact, it takes approximately ($\sim 2$ $s$) for the device’s capacitance to fully decay (Supplementary Figure S4). It is also important to note that the device is slightly stochastic with native cycle-to-cycle variations (Supplementary Figure S5). The cycle-to-cycle variation can also be observed in Figure 1B in the main text in the subtle difference in magnitude between the first recorded capacitance and restored state capacitance. This relatively short-term memory loss in an unstimulated device implies short-term memory and device volatility \cite{Yang2013MemristiveComputing}. The device exhibits pinched hysteresis loops for sinusoidal voltage inputs as observed in its $C-v$, $A-v$, and $W-v$ curves (Supplementary Figure S2).

\maketitle

\newpage
\begin{figure}
    \centering
    \includegraphics[width=4.5 in]{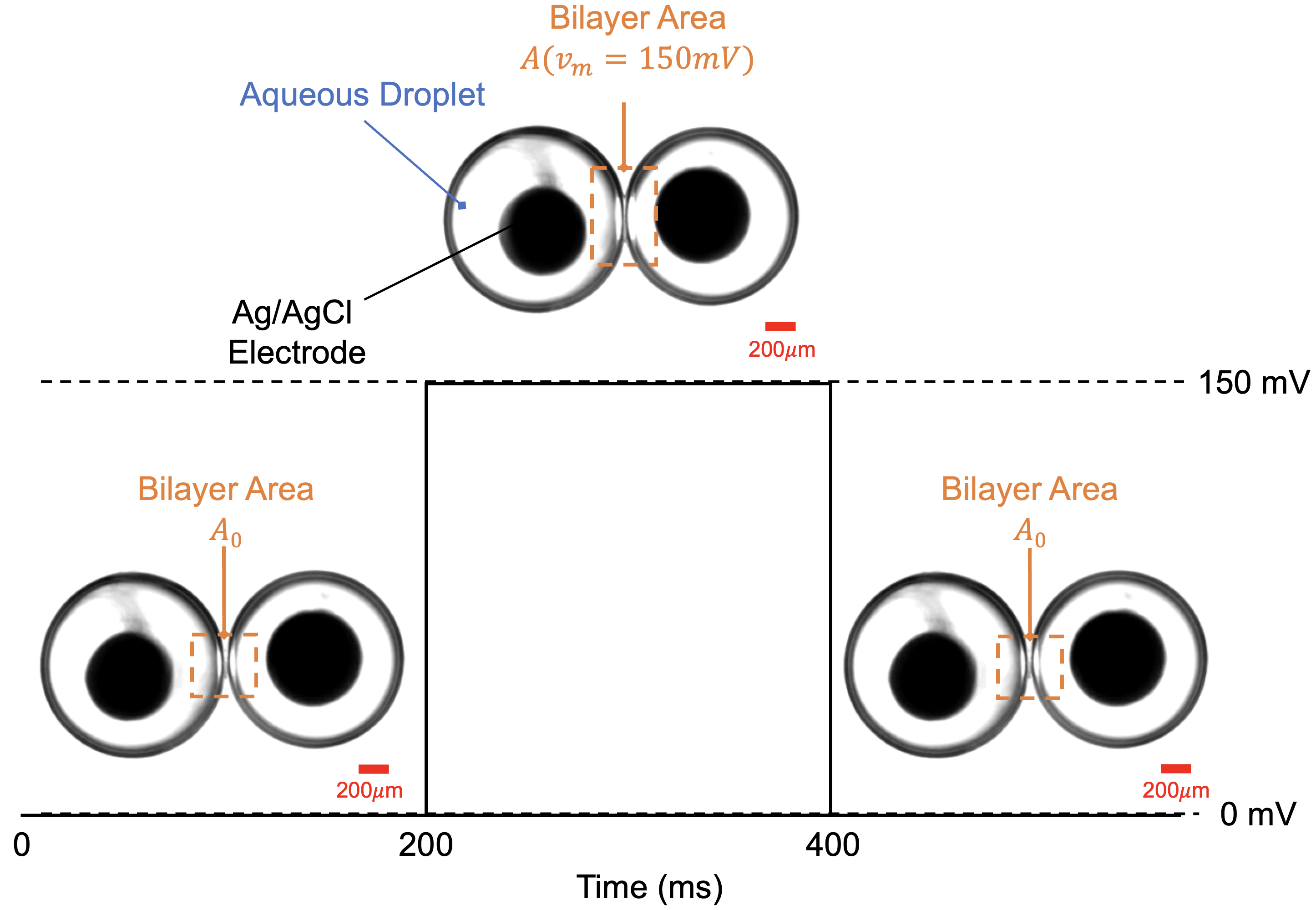}
    \caption{Bottom-view of the memcapacitor in rest and under voltage application. In this figure, we show the memcapacitor's static response, where an increase in transmembrane potential leads to a rise in capacitance and area. For the first 200 $ms$, the memcapacitor is initially unstimulated ($0$ $mV$ applied). In this case, the memcapacitor's bilayer area, at the interface of the two droplets, is observed to be very small, denoted $A_0$. When a 150-$mV$ pulse is applied, the bilayer area increases 2-3 times $A_0$ as observed in the top inset. Upon removal of the stimulus, the device gradually returns to its initial state. }
    \label{RealDeviceSnapshots}
\end{figure}
\begin{itemize}
\color{white}
\item 
\end{itemize}


\newpage
\begin{figure}
    \centering
    \includegraphics[width=4.5 in]{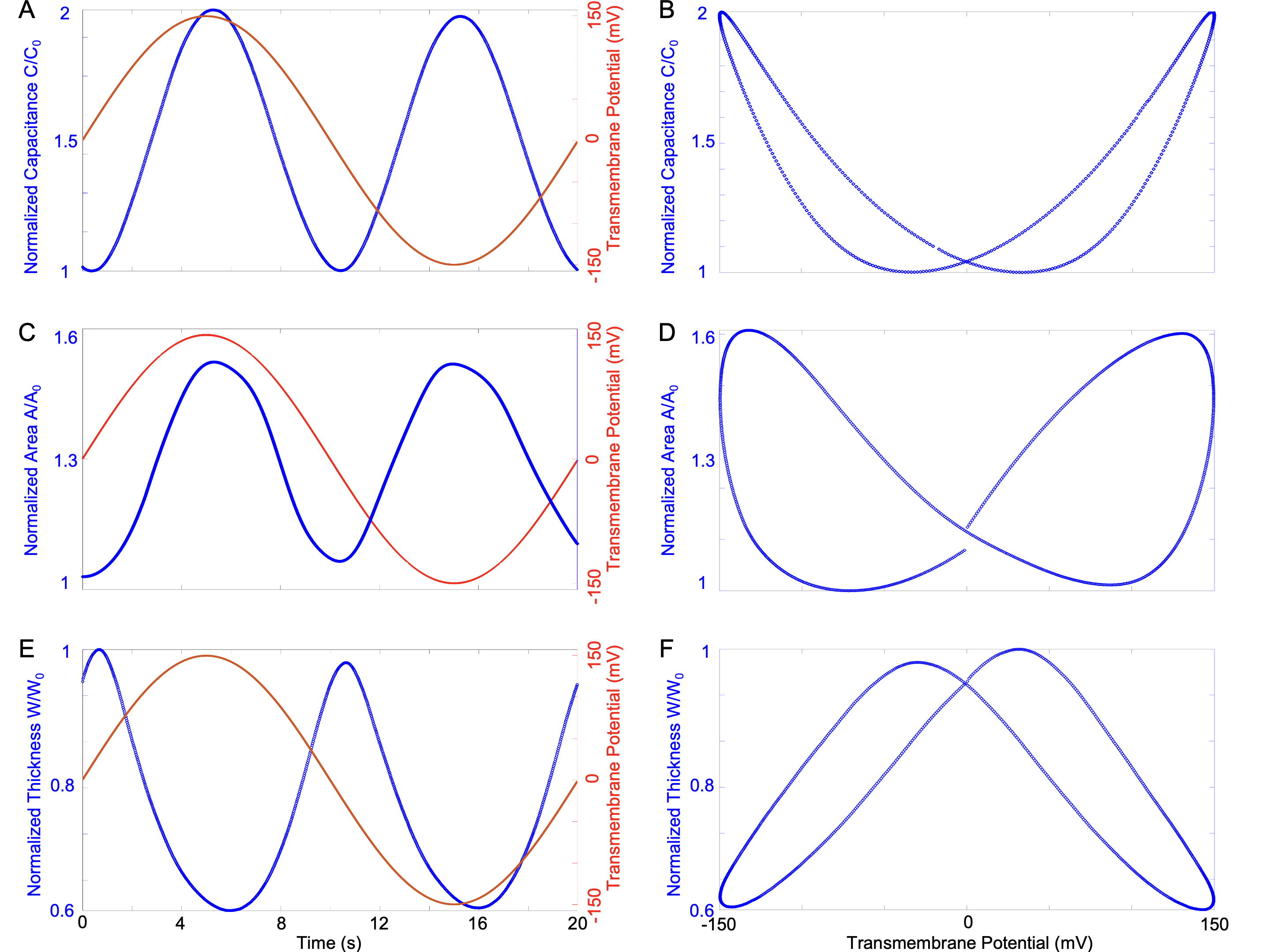}
    \caption{ The memcapacitor exhibits pinched hysteresis loops for its capacitance, bilayer interfacial area (EW), and hydrophobic thickness (EC). (A) Normalized dynamic capacitance as a function of a 150-$mV$, 50-$mHz$, sinusoidal transmembrane potential. (B) Corresponding $C-v$ curve, where a pinched hysteresis observed near 0 $mV$ transmembrane potential. (C) Normalized dynamic bilayer area as a function of a 150-$mV$, 50-$mHz$, sinusoidal transmembrane potential. (D) Corresponding $A-v$ curve, where a pinched hysteresis observed near 0 $mV$ transmembrane potential. (E) Normalized dynamic hydrophobic thickness as a function of a 150-$mV$, 50-$mHz$, sinusoidal transmembrane potential. (F) Corresponding $W-v$ curve, where a pinched hysteresis observed near 0 $mV$ transmembrane potential. For all loops shown, the memcapacitor's time constant for the increasing-voltage path is different from that of the decreasing-voltage path. That is, the device's EW and EC processes exhibit non-reversibility as functions of sinusoidal transmembrane potential, which is a fingerprint of memory. The procedures used to obtain these plots are outlined in the Methods section of the main paper. In addition, pinched $Q-v$ for this device can be found in previous work \cite{Najem2019DynamicalMembranes}.}
    \label{HysteresisLoops}
\end{figure}

\newpage
\begin{figure}
    \centering
    \includegraphics[width=4.5 in]{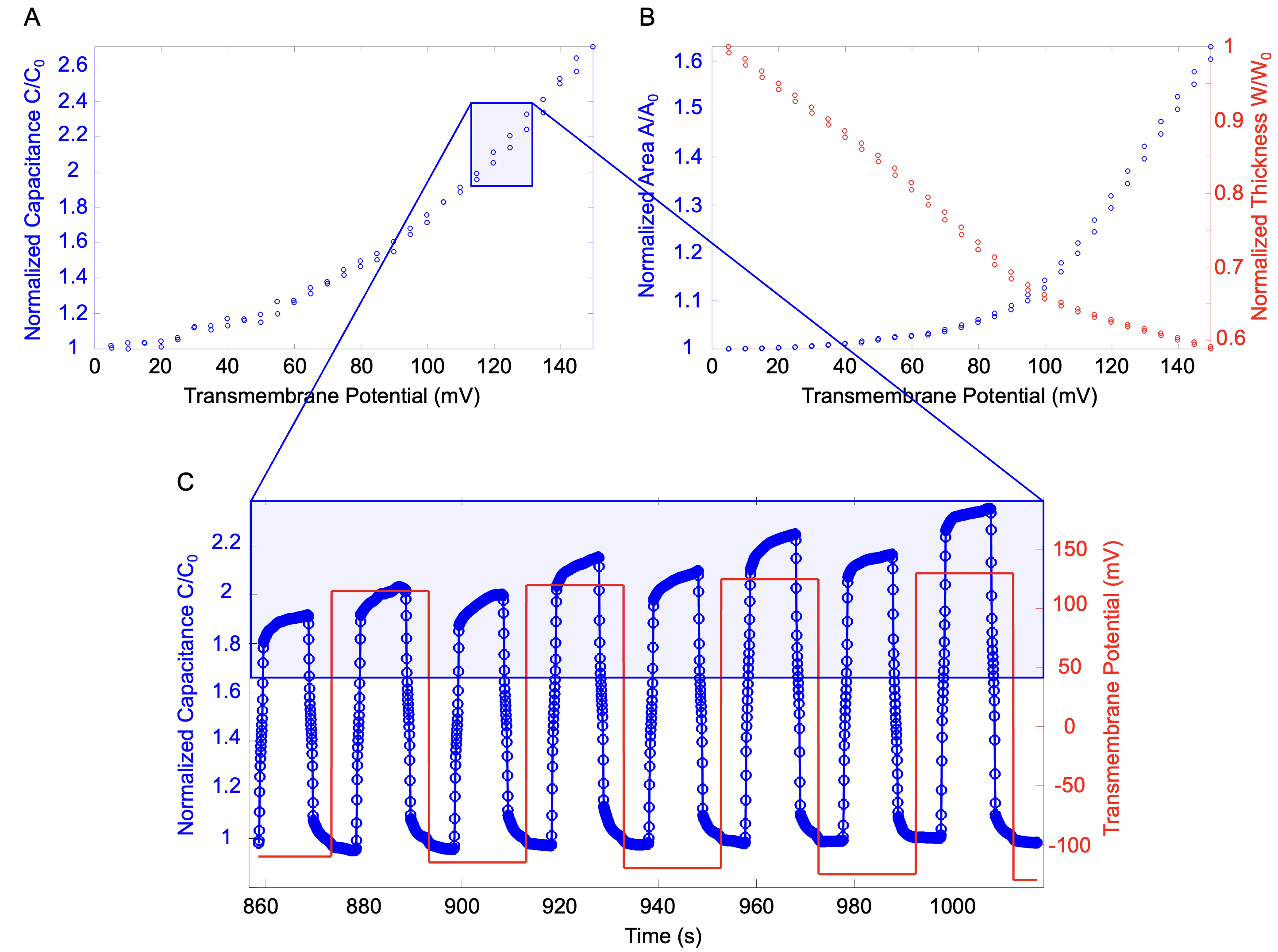}
    \caption{Steady-state responses of capacitance, bilayer area, and hydrophobic thickness. (A) Normalized steady-state capacitance as a function of applied voltage. (B) Left vertical axis: Normalized steady-state interfacial area as a function of applied voltage. Right vertical axis: Normalized steady-state hydrophobic thickness as a function of applied voltage. (C) Left vertical axis: Normalized capacitance step response to an applied step voltage as a function of time. Right vertical axis: Applied train of step voltages as a function of time. For each point in (A), a step input was applied to the memcapacitor for 10 seconds (to ensure the device reaches steady-state), as shown in (C), and the corresponding step response capacitance was computed (refer to the methods section in the main paper) and normalized to the capacitance recorded at rest ($C_0$). The resulting steady-state capacitance values in (A) are obtained by taking the maximum point in the response to the sequential step voltages of amplitudes. For instance, the normalized capacitance values in response to voltages between $\lvert115 $mV$\rvert$ and $\lvert 130 $mV$\rvert$  are shown in (C), where the maximum points of the capacitance responses to the eight shown pulses (-115 $mV$,115 $mV$, -120 $mV$,120 $mV$, -125 $mV$,125 $mV$, -130 $mV$, and 130 $mV$) are obtained and plotted in the blue rectangle in (A). The step responses were implemented in increments of 5 $mV$ for both positive and negative amplitudes, hence two points are observed for each voltage point. For area calculations, we utilized a custom MATLAB image-processing script to estimate the bilayer interfacial area from bottom-view videos of the device under varying step inputs as a function of time. The steady-state values of the normalized interfacial area were obtained by using an identical technique to that shown for the normalized capacitance in (C). Finally, the hydrophobic thickness was computed using equation (3) in the main paper. This method of approximating the hydrophobic thickness as a function of time was developed in earlier work \cite{Taylor2015DirectBilayer}. The steady-state values of the normalized hydrophobic thickness were obtained by using an identical technique to that shown for the normalized capacitance in (C).
    These figures underscore the nonlinearity associated with both EW and EC processes, constituting the nonlinearity in capacitance.}
    \label{SteadyStateResponses}
\end{figure}
    
\begin{itemize}
\color{white}
\item 
\end{itemize}

\newpage
\begin{figure}
    \centering
    \includegraphics[width=4.5 in]{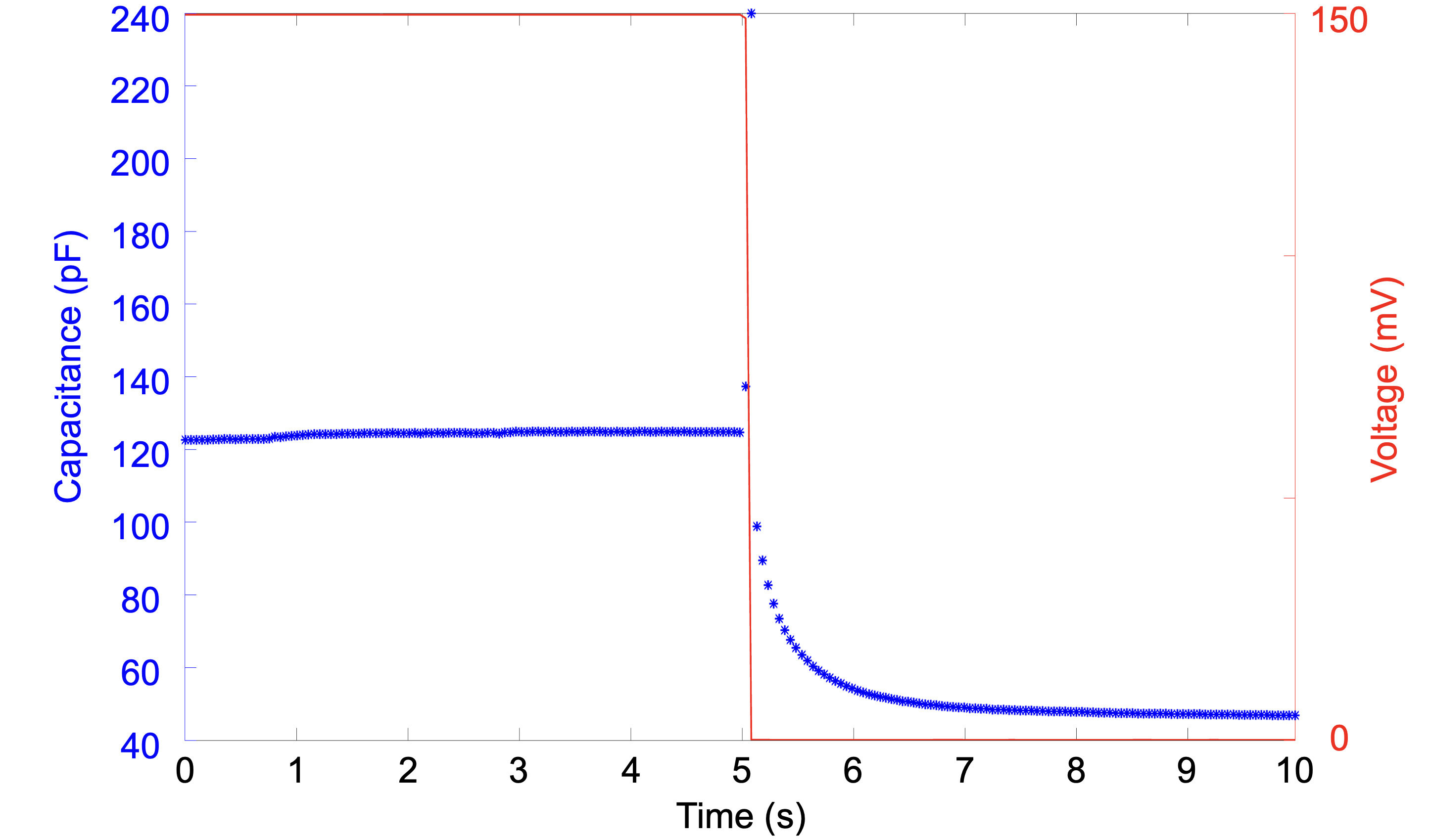}
    \caption{Measured restoration of initial capacitance upon voltage stimulus removal. In this figure, we draw emphasis on the device's fading memory property. The memcapacitance decays in $\sim2$ $s$ when a 150-$mV$ voltage stimulation is removed.}
    \label{FadingMemory}
\end{figure}

\begin{itemize}
\color{white}
\item 
\end{itemize}

\newpage
\begin{figure}
    \centering
    \includegraphics[width=4.5 in]{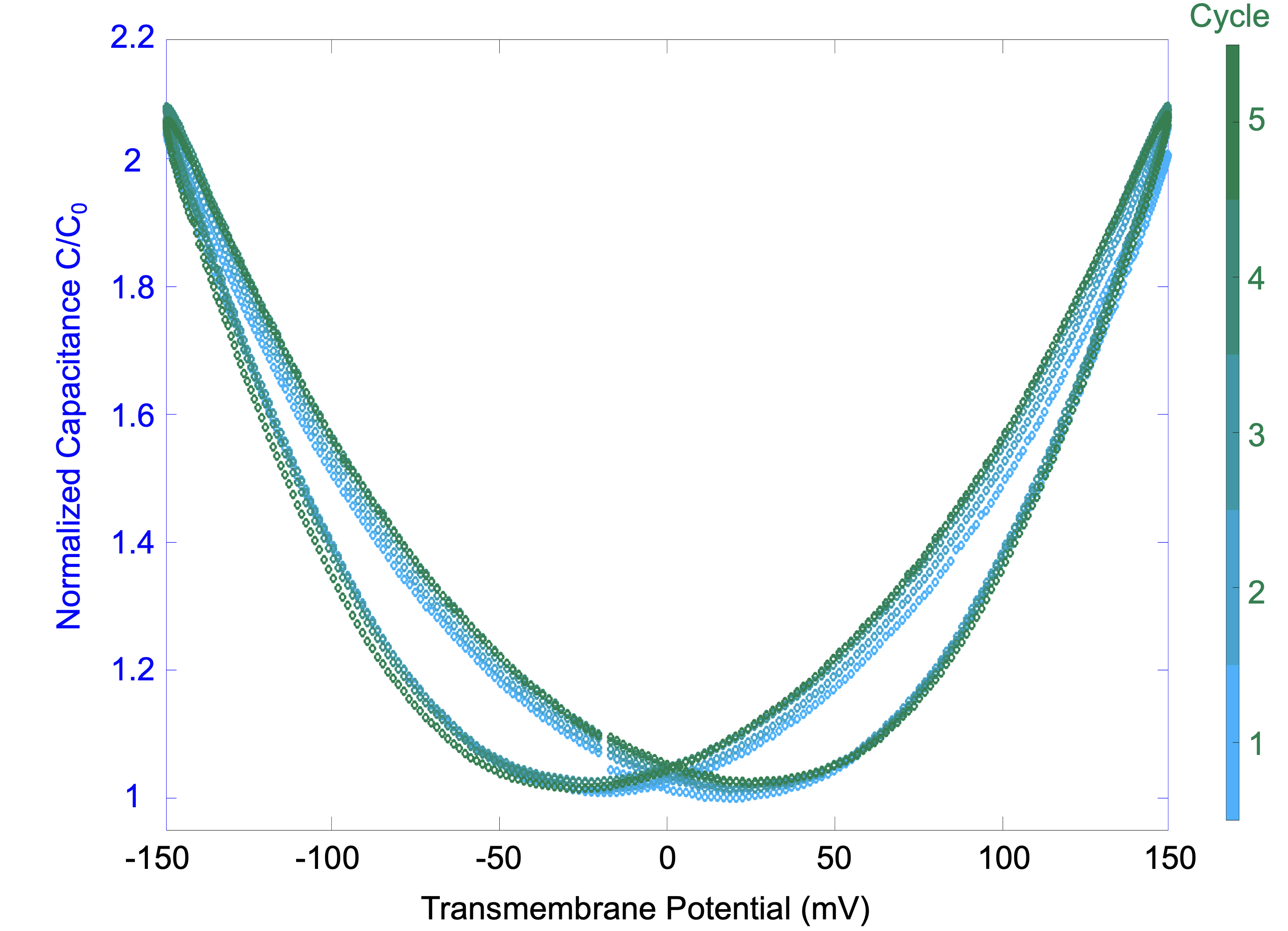}
    \caption{Five cycles of normalized $C-v$ curves. In this figure, we highlight the device's native stochasticity as observed from the $C-v$ curve cycle-to-cycle variation for five cycles of the same applied sinusoidal input (150 $mV$, 50 $mHz$). The exact mechanism behind stochasticity in EW and EC is still not very well understood and beyond the scope of this work.}
    \label{CycleToCycleVariation}
\end{figure}

\begin{itemize}
\color{white}
\item 
\end{itemize}

\newpage
\section{Spoken Digit Classification}\label{sec2}

\begin{figure} [h]
    \centering
    \includegraphics[width=4.5 in]{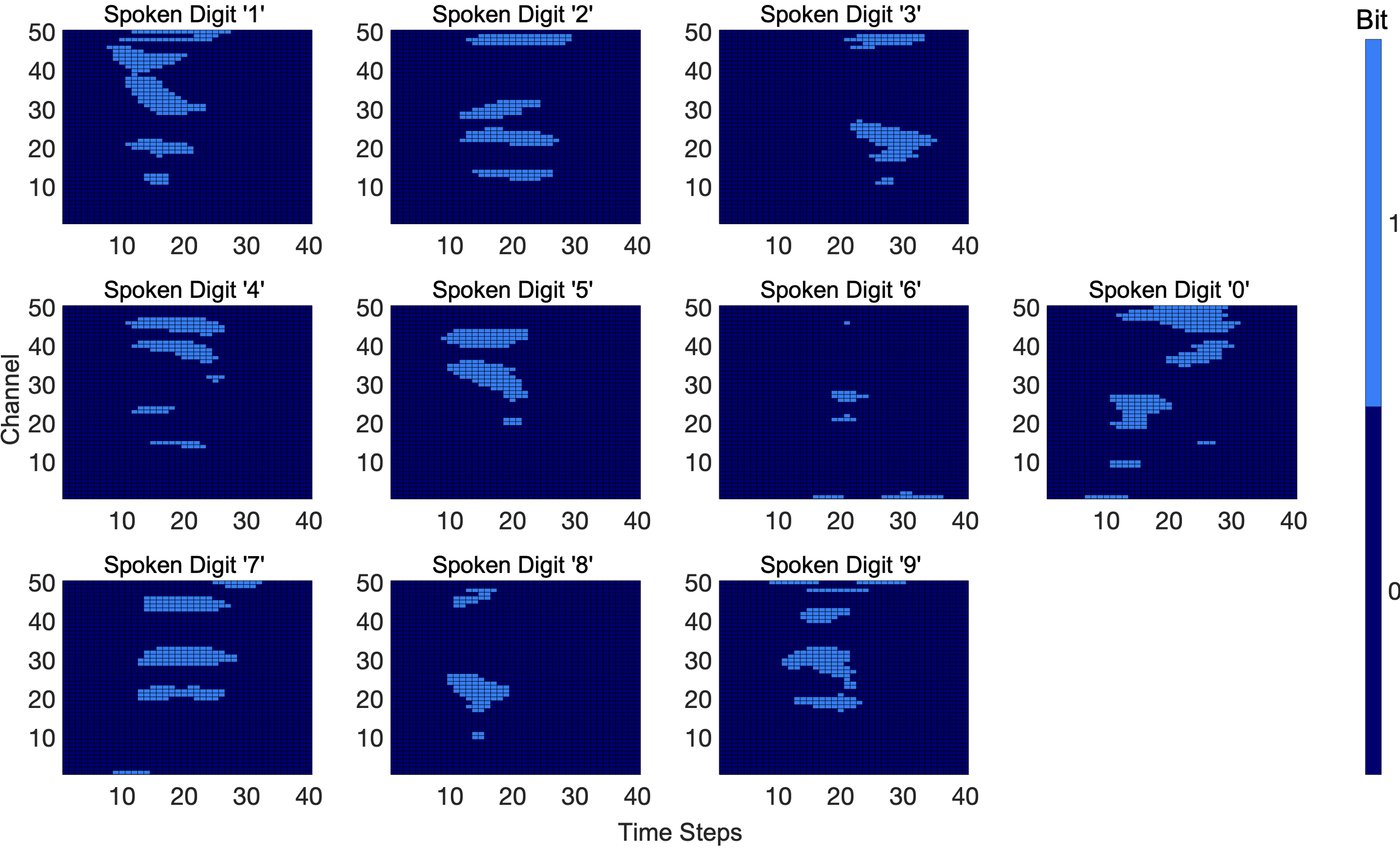}
    \caption{Examples of binary cochleograms for spoken-digits 0-9 (in spoken English).}
    \label{cochleograms}
\end{figure}

\begin{itemize}
\color{white}
\item 
\end{itemize}

\newpage
\begin{figure}
    \centering
    \includegraphics[width=4.5 in]{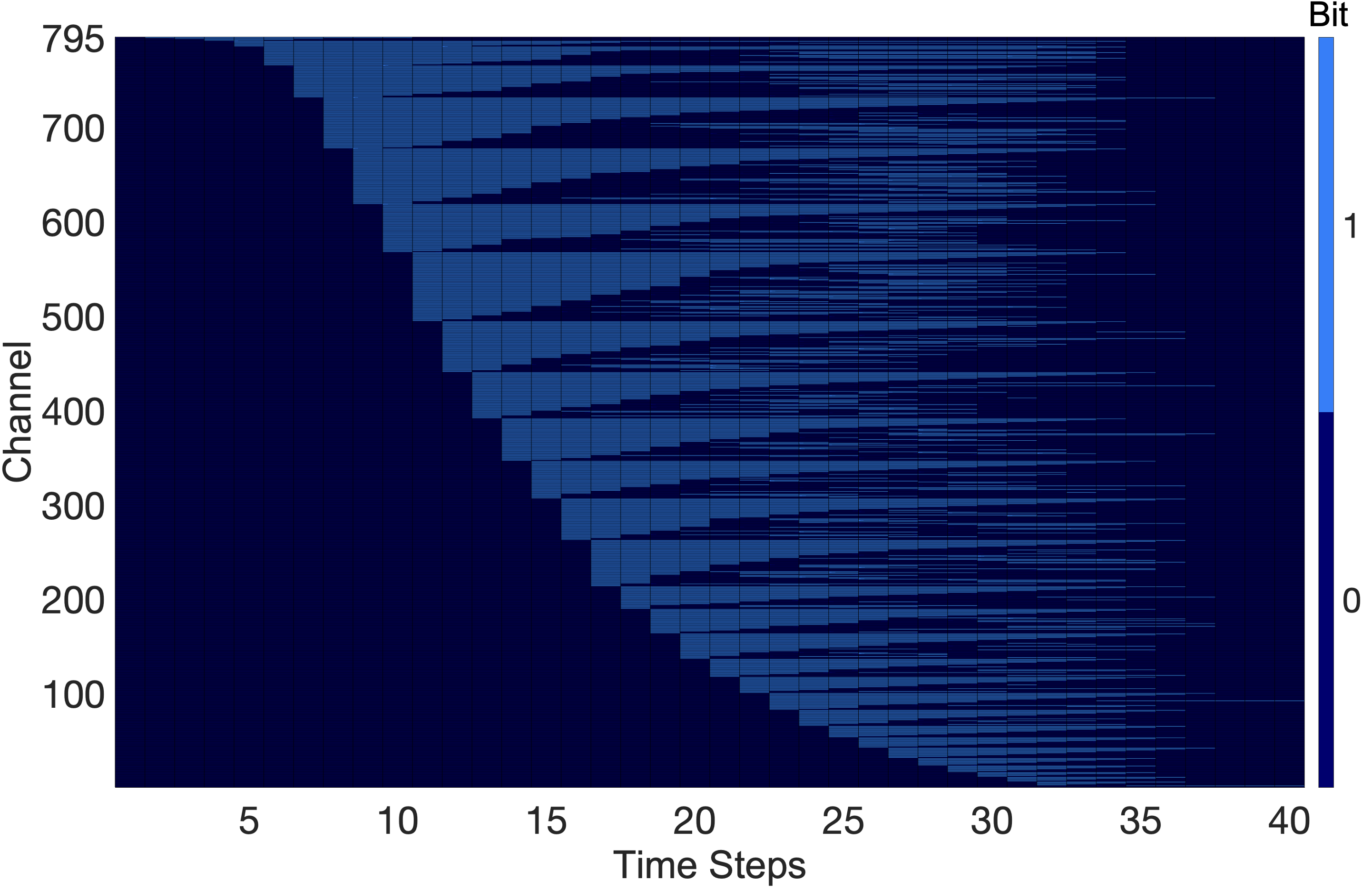}
    \caption{A cochleogram comprising channels with only unique bitstreams across all 500 cochleogram examples used for solving the spoken digit problem. Since encoding all 500 datasets with most channels being populated with 0-bitstreams is redundant, we encoded only the channels with unique bitstream combinations as depicted in Figure 2 in the main text. Encoding only unique channels reduced the number of channels from 25,000 (50 channels for each of the 500 examples) to 795.}
    \label{UniqueCochleogram}
\end{figure}

\begin{itemize}
\color{white}
\item 
\end{itemize}

\newpage
\begin{figure}
    \centering
    \includegraphics[width=4.5 in]{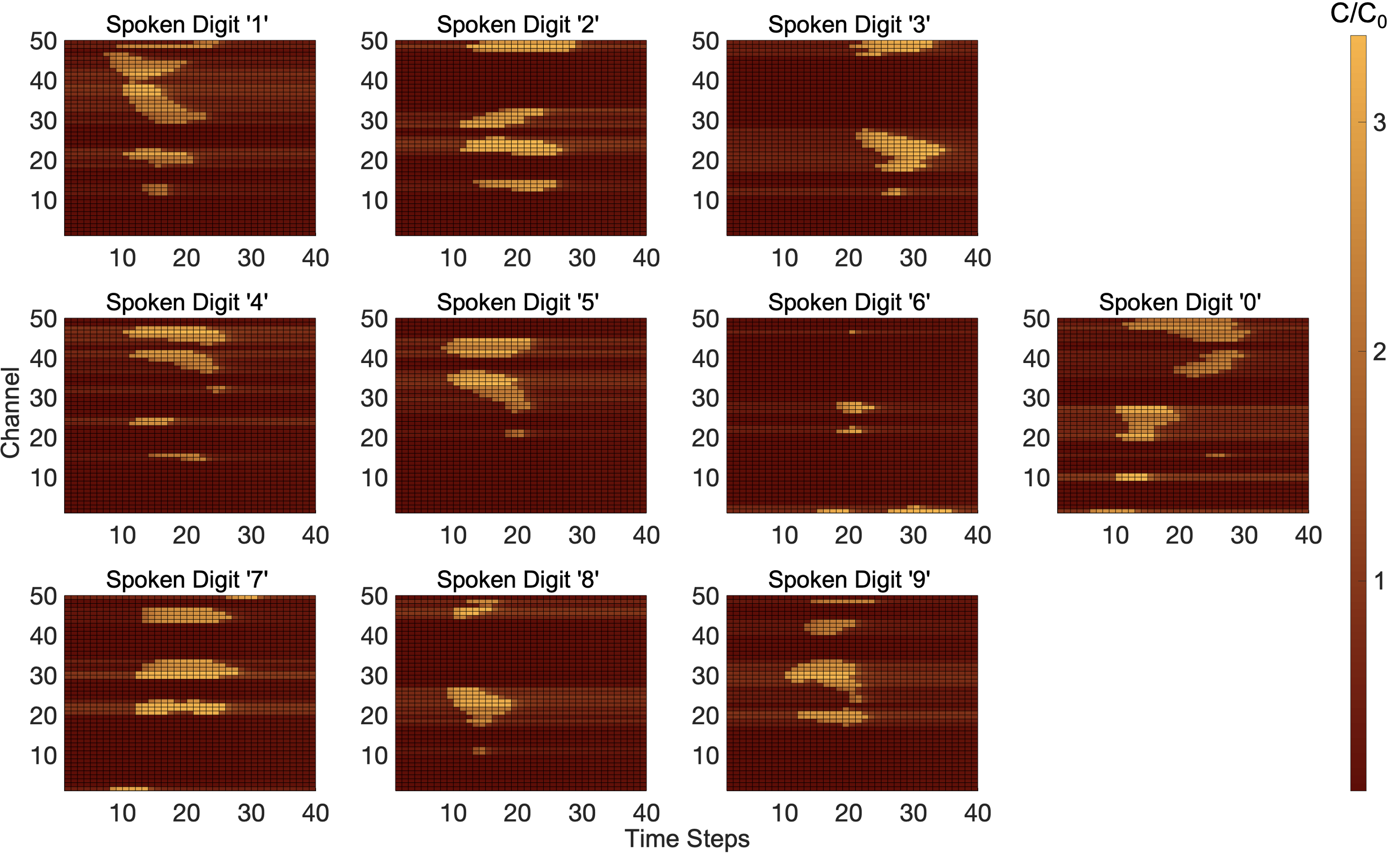}
    \caption{2-D maps depicting the memcapacitor states (normalized capacitance) resulting from voltage stimuli from the example digits shown in Figure S6. The capacitance resulting from the input pulse stream (Figure 2 in the main text) was computed and mapped back into a 2-D array as shown in this figure. Each 2-D map here acts as a visual summary of the device's response to a spoken digit (spoken digits 0-9 as labeled). As observed, the device is able to map 1-bit time steps to large increases in capacitance (2-3 times increase shown in brighter orange) while 0-bit time steps cause negligible changes in capacitance (shown in darker orange).}
    \label{cochleogramsResponse}
\end{figure}

\begin{itemize}
\color{white}
\item 
\end{itemize}

\newpage
\begin{figure}
    \centering
    \includegraphics[width=3.2 in]{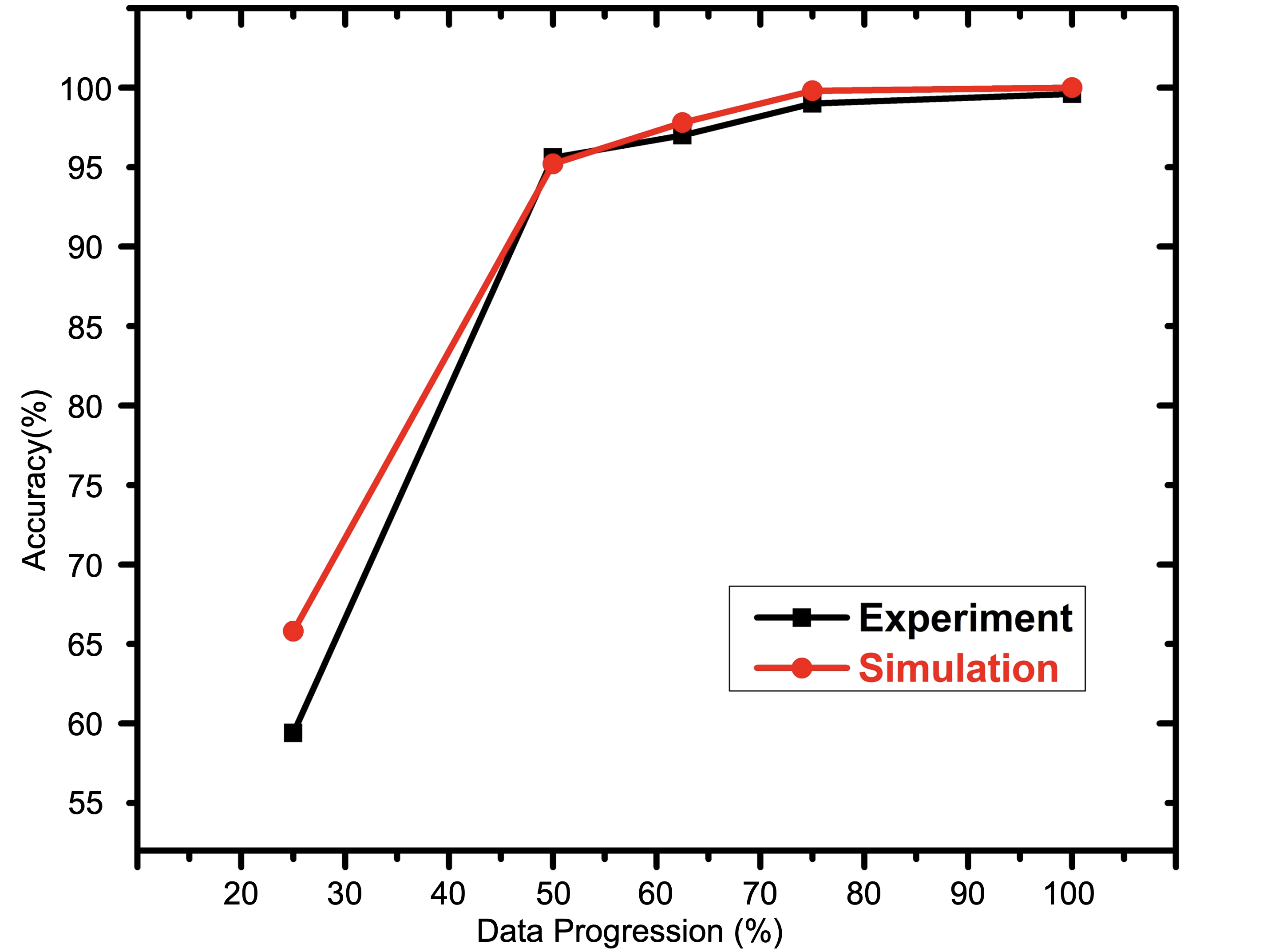}
    \caption{The outcomes for spoken digit classification are displayed in both simulation and experimental settings. The performance is evaluated at various input completion percentages, specifically for $25\%$, $50\%$, $62.5\%$, $75\%$, and $100\%$ (i.e., $10, 20, 25, 30, 40$ time steps) of the test dataset.}
    \label{spoken_result}
\end{figure}
\begin{itemize}
\color{white}
\item 
\end{itemize}

\newpage

\section{Second order nonlinear dynamic task}\label{sec3}

\subsection{Normalized mean squared error}\label{sec4_3}
In this work, we have used the metric normalized mean squared error (NMSE) to calculate the error between the actual and predicted signal, which is defined as:

\begin{equation}
NMSE=\tfrac{\sum_i\left({z_i(t)- y_i(t)}\right)^2}{\sum_i{y_i}^{2}(t)}
\label{NMSE_Du}
\end{equation}
where $z(t)$ is the predicted signal and $y(t)$ is the actual signal. As the actual signal normalizes the result, the error is unitless.

So far, we have computed the NMSE using the previous equation to facilitate equitable comparisons with the work of another group \cite{Du2017ReservoirProcessing}. Moreover, certain studies have placed emphasis on employing the following equation to determine the NMSE value.

\begin{equation}
NMSE=\tfrac{\sum_i\left({z_i(t)- y_i(t)}\right)^2}{\sum_i({y_i(t)- \bar{y}(t)})^{2}}
\label{new_eqn}
\end{equation}

The result of solving the second-order nonlinear dynamic task with new formula has been presented in table \ref{nmse_new}.

\newpage

\subsection{Comparison with a conventional linear network}\label{sec4_2}

Herein, we show a comparison between the memristor-based RC network and the conventional linear network to convey the impact of the intrinsic nonlinear physics of the memristor device. In this section, we have replaced the memristor reservoir layer with a linear hidden layer, which generates 50 random signals of the original input abiding  by the following equation:

\begin{equation}
x(k)= rand(1,100)*u(k)
\end{equation}

where $x(k)$ is the scaled output from the linear resistor instead of the memristor, and $u(k)$ is the input vector. In this particular case, no nonlinear transformation is done by the reservoir. 

\par Figure S11 shows the conventional linear network performance plot for solving a second-order nonlinear dynamic task. The signal fit is not promising, and the error level is higher, especially in the testing dataset. The NMSE values of training and testing datasets are $0.0040$ and $0.0043$, respectively using equation \ref{NMSE_Du}. Therefore, it is conspicuous from the comparison that the intrinsic nonlinear dynamics of the memcapacitor device are important to perform the nonlinear transformation. A summary of the calculated NMSEs can be found in table \ref{nmse_new} using the correct form of NMSE expressed in equation \ref{new_eqn}.

\newpage
\begin{figure}
    \centering
    \includegraphics[width=4.3 in]{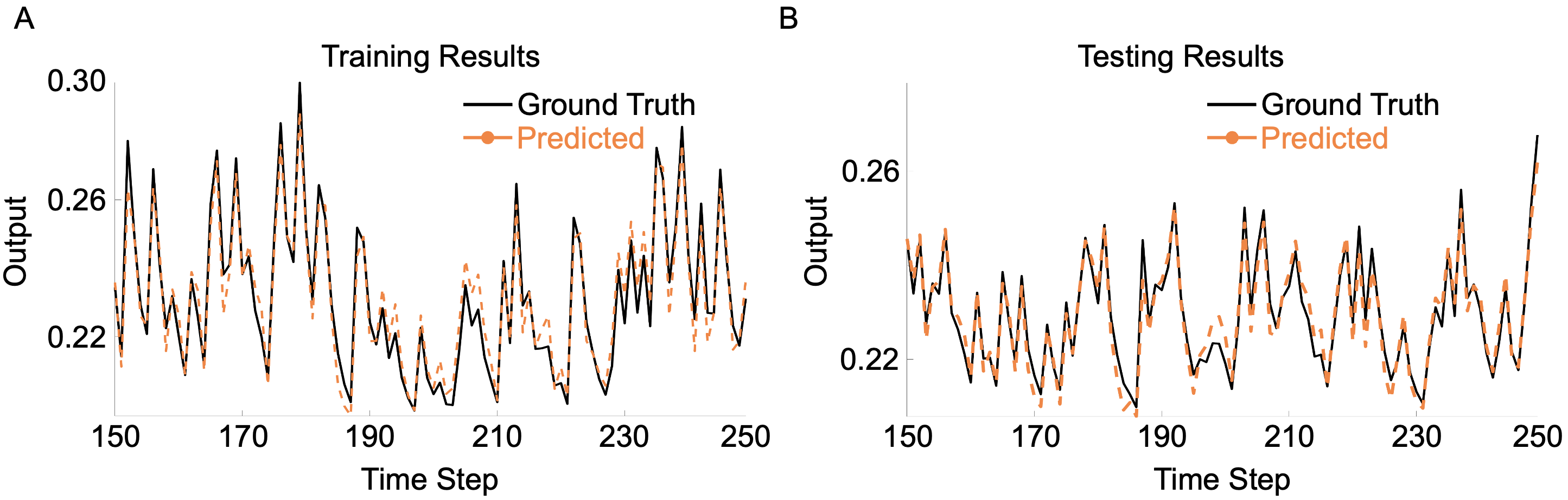}%
    \hspace{1em}%
  
    \caption{\small{Predicting a second-order nonlinear dynamic task using memcapacitor-based RC system in simulation. Ground Truth vs. predicted signals for both (A) training and (B) testing datasets. Our simulation framework is similar to the experimental framework discussed in the main paper. We obtained the capacitance values using the memcapacitor model \cite{Najem2019DynamicalMembranes}. We have achieved an NMSE of $5.28\times 10^{-4}$ and $6.32\times 10^{-4}$ for training and testing data, respectively.}}
    
    \label{SONDS}
  
\end{figure}

\begin{itemize}
\color{white}
\item 
\end{itemize}

\newpage
\begin{figure}[h]
\centering
 \includegraphics[width=4.3 in]{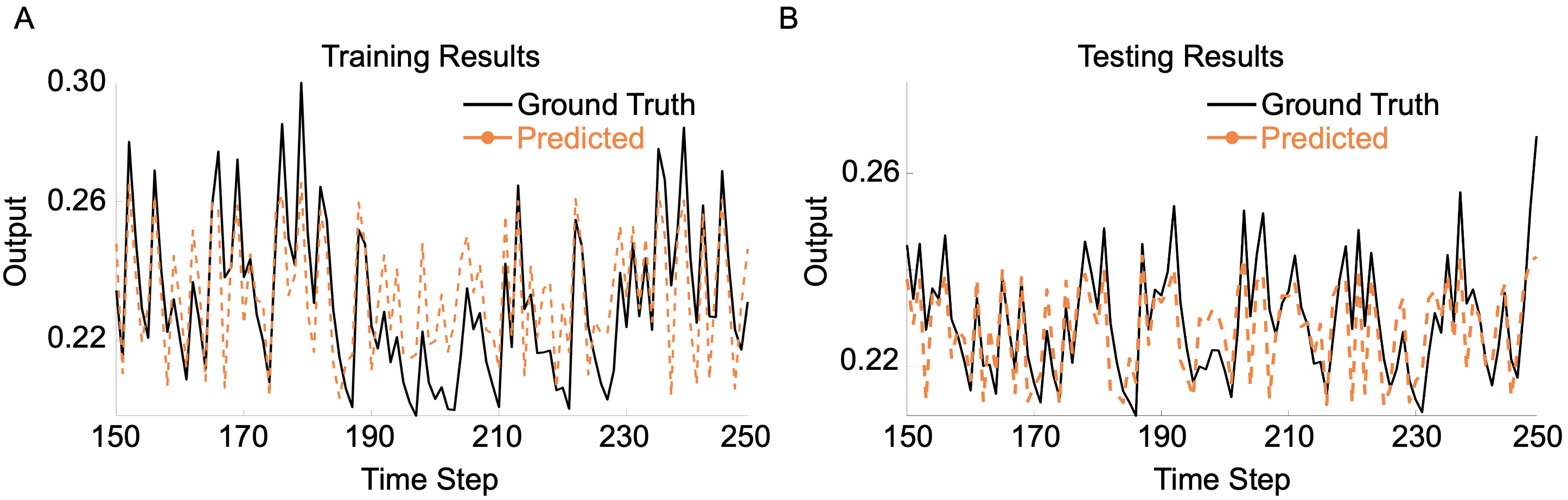}%
 \hspace{1em}%
  
  \caption{\small{Simulating the prediction of a second-order nonlinear dynamic task using the linear network. Ground Truth vs. predicted signals for both (A) training and (B) testing datasets.  We have achieved an NMSE of $0.0040$ and $0.0043$ for training and testing datasets, respectively.}}
  \label{SONDS_LinearNetwork}
  
\end{figure}

\newpage
\begin{table}[h]
\caption{The result of solving second-order nonlinear dynamic task. NMSE was calculated using equation \ref{new_eqn}.}
\centering
\begin{tabular}{|c|c|c|}
\hline
\textbf{Tasks} & \textbf{\begin{tabular}[c]{@{}c@{}}Train\\ (NMSE)\end{tabular}} & \textbf{\begin{tabular}[c]{@{}c@{}}Test\\ (NMSE)\end{tabular}} \\ \hline
Simulation framework   & 0.0526 & 0.0603 \\ \hline
Experimental framework & 0.0573 & 0.0744 \\ \hline
Linear network         & 0.3970 & 0.4067 \\ \hline
\end{tabular}
\label{nmse_new}
\end{table}

\clearpage

\newpage
\section{Real-time Epilepsy Detection from EEG Signal}\label{sec4}
A list of works with testing accuracy for the classification between healthy and epileptic signals has been presented in table \ref{eeg_table} for performance comparison. 
\begin{table}[h]
\caption{Comparison of the proposed method's effectiveness with alternative approaches for differentiating between healthy and epileptic EEG signals. }
\centering
\begin{tabular}{|l|l|c|}
\hline
\multicolumn{1}{|c|}{\textbf{Description}} &
  \multicolumn{1}{c|}{\textbf{Framework}} &
  \textbf{Testing  Accuracy, \%} \\ \hline
\begin{tabular}[c]{@{}l@{}}Memcapacitor based RC system \\ (virtual node)\end{tabular} &
  Simulation &
  99.25 \\ \hline
\begin{tabular}[c]{@{}l@{}}Memcapacitor based RC system\\ (virtual node incorporated with feature modification)\end{tabular} &
  Simulation &
  100 \\ \hline
\begin{tabular}[c]{@{}l@{}}Memcapacitor based RC system \\ (virtual node)\end{tabular} &
  Experiment &
  97.5 \\ \hline
\begin{tabular}[c]{@{}l@{}}Memcapacitor based RC system\\ (virtual node incorporated with feature modification)\end{tabular} &
  Experiment &
  100 \\ \hline
\begin{tabular}[c]{@{}l@{}}Memristor based RC system \cite{Moon2019TemporalSystem} \\ (virtual node)\end{tabular} &
  Simulation &
  99 \\ \hline
\begin{tabular}[c]{@{}l@{}}Memristor based RC system \cite{Moon2019TemporalSystem}\\ (virtual node incorporated with feature modification)\end{tabular} &
  Simulation &
  100 \\ \hline
\begin{tabular}[c]{@{}l@{}}Evolutionary Optimization for Neuromorphic 
\\System (EONS) \cite{Schuman2016AnArchitectures}\end{tabular}&
  Simulation &
  98.25 \\ \hline
\end{tabular}
\label{eeg_table}
\end{table}

\begin{itemize}
\color{white}
\item 
\end{itemize}
\newpage
\section{IRIS dataset classification}\label{sec5}

\par This work demonstrates an approach to classify the Iris dataset using a memcapacitive reservoir system. The Iris dataset is a well-known static data benchmark in machine learning, consisting of 150 samples of four features (petal length, petal width, sepal length, and sepal width) and three species (Iris Setosa, Iris Versicolour, and Iris Virginica) \cite{Jeong2018K-meansNetworks}. The aim of this work is to demonstrate the capability of our system in classifying static datasets. The Iris dataset comprises 150 samples, with 50 samples for each species, and has been divided into a training set (60\%) and a testing set (40\%). The data was converted into voltage signals ranging from 100 $mV$ to 200 $mV$, with a pulse width of 2.5 seconds, and applied to the memcapacitive reservoir system. For each sample, four reservoir states were obtained at a time, and these states were then passed through a $4\times3$ readout layer, which was trained using logistic regression.

\par The results of our experimentation showed a recognition rate of approximately 98.33\%, as demonstrated in Figure S12. This recognition rate is a testament to the effectiveness of our memcapacitive reservoir system in classifying the Iris dataset. This result highlights the potential of memcapacitive reservoir systems in handling complex datasets and further confirms their capability as a promising alternative to traditional machine learning algorithms.

\newpage
\begin{figure}[h]
    \centering
    \includegraphics[width=4.5 in]{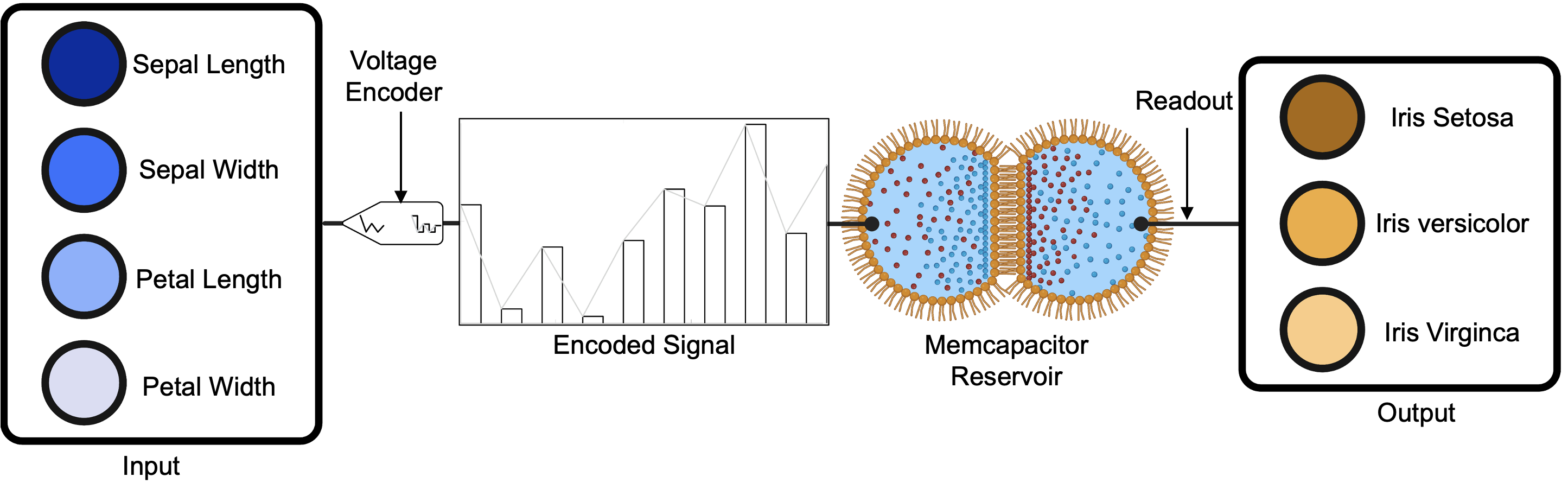}
    \caption{The process flow of Iris Dataset classification using a memcapacitor-based RC system. The input is first encoded into voltage signals ranging from 100 $mV$ to 200 $mV$ with a pulse width of 2.5 seconds. The input is then fed to the memcapacitor model \cite{Najem2019DynamicalMembranes}. The output state matrix is then used for training the readout layer between the memcapacitor and the output layer.}
    \label{Iris}
\end{figure}
\clearpage